\documentclass[sigplan,10pt]{acmart}
\AtBeginDocument{%
  }

\usepackage{tikz}
\usepackage{amsmath}
\usepackage{mathtools}
\usepackage{empheq}
\usepackage{cleveref}
\usepackage{xspace}
\usepackage{bbm}
\newcommand{\states}{\mathcal{S}}
\newcommand{\actions}{\mathcal{A}}

\newcommand{\lcc}{LCC\xspace}
\newcommand{\lccs}{LCCs\xspace}

\newcommand{\Reals}{\mathbb{R}}
\newcommand{\conc}{\gamma}

\newcommand{\cwnd}{\textsc{cwnd}\xspace}

\newcommand{\mdp}{\mathcal{M}}

\newcommand{\expec}{\mathop{\mathbb{E}}}

\newcommand{\del}{\textsc{delay}\xspace}
\newcommand{\loss}{\textsc{loss}\xspace}
\newcommand{\thr}{\textsc{thr}\xspace}
\newcommand{\tcp}{\textsc{TCP}\xspace}
\newcommand{\vol}{\textsc{Volume}\xspace}

\newcommand{\certificate}{\textsc{QC}\xspace}
\newcommand{\certificates}{\textsc{QC}s\xspace}
\newcommand{\qc}{\textsc{QC}\xspace}

\newcommand{\qcf}{\textsc{QC}$_\text{feedback}$\xspace}
\newcommand{\qcsat}{\textsc{QC}$_\text{sat}$\xspace}

\usepackage{subcaption}
\usepackage{multirow}
\usepackage{makecell}
\usepackage{enumerate}
\usepackage{enumitem}
\usepackage{amsmath}
\usepackage[normalem]{ulem}
\hypersetup{breaklinks=true}

\copyrightyear{2026}
\acmYear{2026}
\setcopyright{cc}
\setcctype{by}
\acmConference[EUROSYS '26]{21st European Conference on Computer Systems}{April 27--30, 2026}{Edinburgh, Scotland Uk}
\acmBooktitle{21st European Conference on Computer Systems (EUROSYS '26), April 27--30, 2026, Edinburgh, Scotland Uk}
\acmPrice{}
\acmDOI{10.1145/3767295.3769362}
\acmISBN{979-8-4007-2212-7/26/04}

\begin{document}
%-------------------------------------------------------------------------------

%don't want date printed
\date{}

% make title bold and 14 pt font (Latex default is non-bold, 16 pt)

\title{Canopy: Property-Driven Learning for Congestion Control}

\author{Chenxi Yang}
\affiliation{%
  \institution{The University of Texas at Austin}
  \city{Austin}
  \state{Texas}
  \country{USA}
}
\author{Divyanshu Saxena}
\affiliation{%
  \institution{The University of Texas at Austin}
  \city{Austin}
  \state{Texas}
  \country{USA}
}
\author{Rohit Dwivedula}
\affiliation{%
  \institution{The University of Texas at Austin}
  \city{Austin}
  \state{Texas}
  \country{USA}
}
\author{Kshiteej Mahajan}
\affiliation{%
  \institution{Google DeepMind}
  \city{Mountain View}
  \state{California}
  \country{USA}
}
\author{Swarat Chaudhuri}
\affiliation{%
  \institution{The University of Texas at Austin}
  \city{Austin}
  \state{Texas}
  \country{USA}
}
\author{Aditya Akella}
\affiliation{%
  \institution{The University of Texas at Austin}
  \city{Austin}
  \state{Texas}
  \country{USA}
}

\renewcommand{\shortauthors}{C. Yang et al.}

\newcommand{\sysname}{\textsc{Canopy}\xspace}

\begin{abstract}

Learning-based congestion controllers offer better adaptability compared to traditional heuristics. However, the unreliability of learning techniques can cause learning-based controllers to behave poorly, creating a need for formal guarantees. While methods for formally verifying learned congestion controllers exist, these methods offer binary feedback that cannot optimize the controller toward better behavior. We improve this state-of-the-art via \sysname, a new \emph{property-driven} framework that integrates learning with formal reasoning in the learning loop. \sysname uses novel quantitative certification with an abstract interpreter to guide the training process, rewarding models, and evaluating robust and safe model performance on worst-case inputs. Our evaluation demonstrates that unlike state-of-the-art learned controllers, \sysname-trained controllers provide both adaptability and worst-case reliability across a range of network conditions. %\chenxi{rephrased. PTAL.}

%-------------------------------------------------------------------------------
% Your abstract text goes here. Just a few facts. Whet our appetites.
% Not more than 200 words, if possible, and preferably closer to 150.
\end{abstract}

%%
%% The code below is generated by the tool at http://dl.acm.org/ccs.cfm.
%% Please copy and paste the code instead of the example below.
%%


%%
%% Keywords. The author(s) should pick words that accurately describe
%% the work being presented. Separate the keywords with commas.

\maketitle
\def\thefootnote{*}\footnotetext{Saxena and Dwivedula contributed equally to this work.}\def\thefootnote{\arabic{footnote}}

%-------------------------------------------------------------------------------

\section{Introduction}
\label{sec:introduction}
Effectively controlling congestion in networks is a longstanding problem. Several recent approaches to it use neural models, trained using reinforcement learning (RL). Such neural models have been shown to significantly outperform classical controllers in a range of network environments~\cite{abbasloo2020classic,orca-followup2023,jay2019deep,vivace2018nsdi}.

However, the unreliability of these neural models seriously impedes 
their deployment in real-world performance-critical networks today. Indeed, as we demonstrate in \Cref{sec:motivation}, neural congestion controllers can exhibit highly suboptimal behavior, failing to meet key {\em properties} on worst-case inputs. For example, they can take {\em aggressive, poor actions in response to noise} in network state measurements; and their {\em congestion window (\cwnd) adjustments can be completely misaligned with current network conditions}, such as drastically lowering  \cwnd even when network conditions are good.

The question, then, is: how do we improve property satisfaction in learned controllers? Prior works that formally verify learned congestion controllers~\cite{eliyahu2021verifying}, while useful, suffer from a key limitation: they offer one-time boolean feedback on whether the controller satisfies a property, and in case verification fails, we have no option but to train an entirely new controller that may fail verification all over again.

In this paper, we take a different approach to learning congestion controllers that achieve high property satisfaction. Our approach, called \sysname, \emph{integrates learning with in-the-loop symbolic analysis of worst-case behavior}. In \sysname, developers of congestion control algorithms specify: (i) a backbone learning-based algorithm with a reward function that reflects an average-case performance objective, and (ii) a set of {\em properties that the controller should satisfy even on worst-case inputs}. These properties reflect designers' conventional wisdom on ``best to avoid" congestion control behaviors. %\aditya{why not also have "good to have" or "best to avoid"? we don't need to be too precise here} \chenxi{I am slightly leaning towards not having 'good to have' to be consistent with properties in later sections.}

\sysname's analysis procedures are based on {abstract interpretation} \cite{cousot1977abstract}, a classic framework for sound formal verification of systems. We observe that a 100\% property satisfaction is hard to achieve for a learning algorithm: such a model may not exist, and even if it did, searching for the global optimum model under constraints imposed by the properties is computationally difficult for complex, high-dimensional systems.
Therefore, we propose a new object in our analysis procedure, which we call a {\em quantitative certificate} (\certificate). 
The \qc captures: (A) {\em feedback}, which informally 
measures the relative volume of an input region that provably satisfies a given property (a refined definition is presented below), 
and (B) a proof of property satisfaction for that region.  The learning algorithm in \sysname uses the feedback from the \certificate ((A) above) as part of the training loss. In effect, the \sysname approach leverages property-driven \certificates to (1) impose restrictions on and regulate a learned congestion controller by guiding it on what (not) to
do and (2) ensure that training iterations drive the learned controller towards \emph{high worst-case satisfaction of properties} \underline{and} high average-case performance. Further, \certificates extend beyond \sysname's training phase — they can also serve as runtime features to quantify and monitor a controller’s behavior.

We instantiate the \sysname framework to Orca, a recently-proposed RL-based congestion controller~\cite{abbasloo2020classic}. During training, we use abstract interpretation to compute symbolic overapproximations of the set of all possible mappings between Orca's inputs (network states) and outputs (\cwnd). During each training step, the symbolic set is split into several components (each overapproximating an input range). The \certificate then helps capture property satisfaction by computing the aggregated distances between the outputs from each input component and the set of outputs that satisfy the property. We use this distance to shape the RL reward function. The outcome is that, at convergence, the symbolic mapping we construct establishes \emph{quantitative bounds on the trained controller's worst-case behavior} and {\em fully describes input regions where the learned algorithm provably satisfies the property}. 

In this paper, we showcase the utility of the \sysname approach for a set of five properties targeting different network conditions and desired outcomes (\Cref{subsec:properties}). Similar to prior works on designing LCCs~\cite{abbasloo2020classic} and verifying them~\cite{eliyahu2021verifying}, we focus on LCC properties for a single flow. We do not claim that these properties are exhaustive. However, by showcasing a diverse set, we aim to demonstrate that CC designers can craft properties tailored to the sophistication of their deployment scenarios and integrate them with \sysname.

We conduct testbed evaluations of \sysname on a set of synthetic and real-world traces, and deploy the controller on a global testbed. 
We find that \sysname models achieve up to 1.4$\times$ higher worst-case satisfaction for key properties than Orca. The \sysname controller reduces delays by up to 61\% compared to Orca and up to 37\% smaller delays compared to TCP Cubic, while maintaining comparable or superior bandwidth utilization. Our real-world deployment on a global testbed demonstrates that \sysname consistently outperforms Orca and TCP Cubic, while preserving comparable bandwidth utilization. We further demonstrate another use case of \certificates where they are used to monitor a controller’s property satisfaction \textit{at runtime} and adjust the controller on the fly. Our results show that \certificates can serve as online monitors and improve performance when integrated with learning-based congestion controllers (LCCs).

In summary, our key contributions are: 
\begin{itemize}[]
    \item We motivate the need to enable learned congestion controllers (and, more generally, RL-controlled software systems) to certifiably satisfy operator-desired properties. 
    \item We introduce quantitative certificates (\certificates) to measure and prove fine-grained property satisfaction. 
    \item We present \sysname a learned congestion control approach that achieves ``certification in the learning loop'' by using \certificates to regulate the learner toward satisfying desired properties.
    \item Our implementation and evaluation of \sysname across synthetic and real-world network traces and in-the-wild experiments show that its models have high worst-case satisfaction for key properties while preserving or improving average-case performance  compared to learned controllers. The prototype of \sysname is available at \url{https://github.com/ldos-project/Canopy}.
\end{itemize}

\section{Motivation and \sysname{} Overview} \label{sec:motivation}

Learning-based congestion controllers (\lccs) \cite{abbasloo2020classic, vivace2018nsdi, jay2019deep, orca-followup2023} have shown great promise in adapting to complex and dynamic network environments and doing well on the average case. However, concern around their behavior {\em in the worst case} is a key factor preventing broad use.
%, they often operate as black boxes. %This sets them apart from traditional congestion control algorithms like TCP Cubic \cite{ha2008cubic}, which are manually designed. 
%As a result, their % opacity of \lccs makes their 
%decision-making processes are difficult to interpret, leaving their worst-case behavior poorly understood. Additionally, studies on their worst-case performance bounds remain sparse, further raising concerns about their reliability in critical scenarios.}
We use Orca~\cite{abbasloo2020classic}\footnote{While there are several RL-based controllers~\cite{abbasloo2020classic, jay2019deep, orca-followup2023} for congestion control and our framework can work with any of them, we chose Orca owing to its available and reproducible codebase.}, an existing \lcc, to illustrate poor behaviors, where the congestion controller violates common-sense properties. % as the basis for analysis in this paper
Then, we sketch our approach to addressing the underlying issues.

%We begin by demonstrating scenarios where a state-of-the-art learned congestion controller (\lcc) behaves suboptimally. Next, we sketch our approach to addressing these issues.

\subsection{Issues with Existing LCCs} \label{subsec:motivation-examples}

Orca is a deep RL agent that monitors network states (such as queuing delay, observed throughput, etc.) and periodically modifies the congestion window (\cwnd) computed by its `backbone' manually designed congestion controller, namely, TCP Cubic (see details in \Cref{subsec:bg-orca}).
Thus, fine-grained control is performed by TCP Cubic and coarse-grained `corrections' are made by Orca's \lcc.

We analyze the behavior of Orca for two desirable properties that congestion control designers may care about: (i) Noise Robustness: i.e., avoiding drastic performance changes when measurements of network statistics are noisy; and (ii) Safety: avoiding significant and persistent bandwidth under- or over-subscription.
% (i) {\bf Robustness:} Can Orca maintain high performance on traces similar to, but with slight differences from, those where it was trained on and has already excelled?; and (ii) {\bf Performance:} Does the Orca controller utilize bandwidth well without significantly under- or over- subscribing?
To this end, we employ several hand-constructed traces with frequent but controlled available bandwidth variations. % and distill our findings into two observations:

\begin{figure}[t]
    \centering
    \begin{subfigure}{\linewidth}
        \centering
        \includegraphics[width=0.8\linewidth]{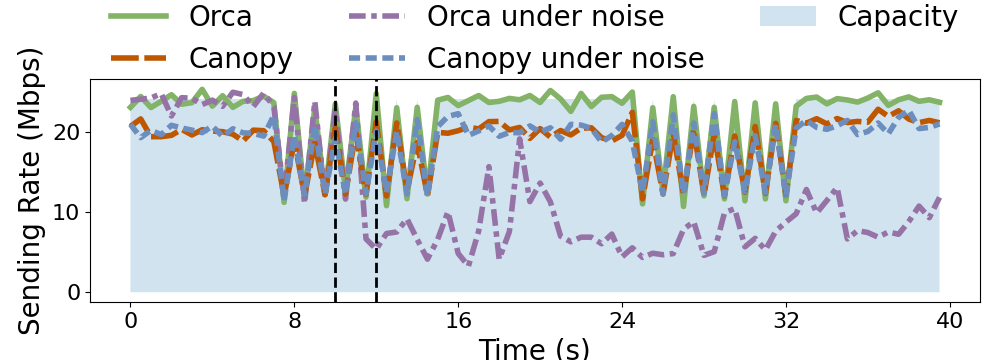}
        \caption{`Orca under noise' and `\sysname under noise' are Orca and \sysname models subjected to {\it at most} 5\% uniformly random noise to the observed queuing delay.
        }
        \label{fig:orca-noises-trace}
    \end{subfigure}
    \begin{subfigure}{\linewidth}
        \centering
        \includegraphics[width=0.8\linewidth]{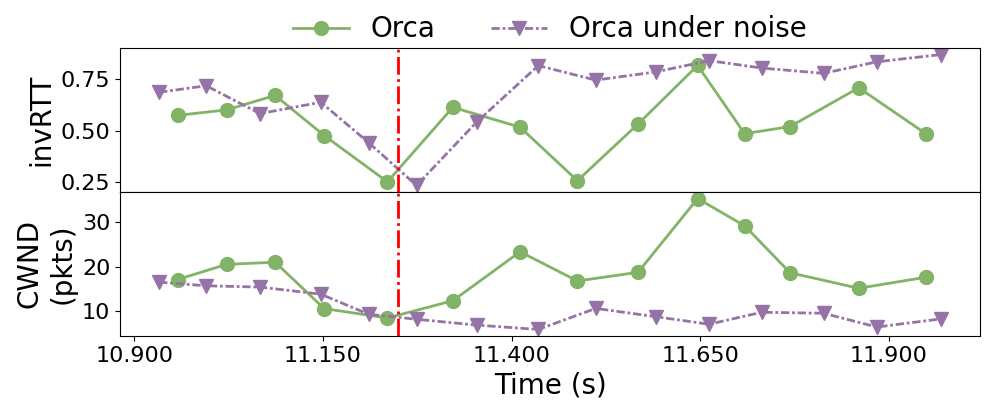}
        \caption{invRTT represents the inverse normalized RTT ($minRTT/RTT$). High invRTT implies close to minimum RTT and hence should be followed by higher congestion windows in future time steps.}
        \label{fig:orca-noises-state}
    \end{subfigure}
    \caption{\small{Orca under noises -- (a) Sending Rate of Orca and \sysname with and without noise. \sysname is much more robust. (b) invRTT observed by Orca (including the noise) and its \cwnd, during the region within the dashed lines in (a). After the dashed vertical line, Orca under noise continues to use a small \cwnd while Orca, with a similar input, uses a higher \cwnd.}}
    \label{fig:orca-noises}
 \vspace{-0.15in}\end{figure}

\noindent
{\bf Observation \#1: Orca is sensitive to random noise.}
Consider the synthetic trace in \Cref{fig:orca-noises}. Orca performs reasonably on this trace, mostly utilizing the available bandwidth even under bandwidth fluctuations. 
We then add {\em random noise}, uniformly sampled between -5\% and 5\%, to the observed delay state before passing it to the Orca \lcc. %\footnote{
%Note that such small perturbations can manifest in practice due to measurement noise.
%The 5\% noise is used for bounding the single step's \cwnd change. 
Such measurement noise is common, e.g., due to the coarse granularity of throughput or drop-rate feature measurements in \lccs like Orca. 
%If we observe consistent 5\% noise, we consider it as a real queuing delay change and would not constrain the accumulative \cwnd change. }
The final state seen by the controller is shown as invRTT in \Cref{fig:orca-noises-state}. We see that
%\Cref{fig:orca-noises-state} shows that 
{\em even when the noisy input invRTT is high (e.g., right before 11.4s) implying low delays, the \lcc continually decreases and maintains a low \cwnd} leading to severe under-utilization! In general, the underlying statistical decision-making of \lccs can lead to arbitrary bad decisions such as above, especially if the controller was not exposed to such conditions during training. %\aditya{say that learned models like orca are blackboxes and their underlying statistical decision making can cause them to make bad decisions such as drastically lowering cwnd when not called for}
%Thus, input points with similar features can lead to very different trajectories with Orca, showing that it is {\em not robust} to small amounts of noise.

\begin{figure}[t]
    \centering
    % First subfigure
    \begin{subfigure}{\linewidth}
        \centering
        \includegraphics[width=0.8\linewidth]{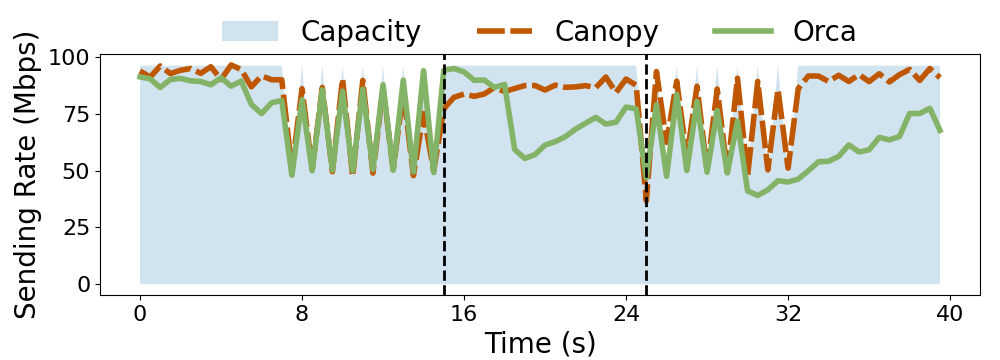}
        \caption{Sending rate of Orca (an \lcc{} without \certificate{}s) vs. \sysname{} (trained with \certificate{}s).}
        \label{fig:orca-critically-bad-trace}
    \end{subfigure}
    % Second subfigure
    \begin{subfigure}{\linewidth}
        \centering
        \includegraphics[width=0.8\linewidth]{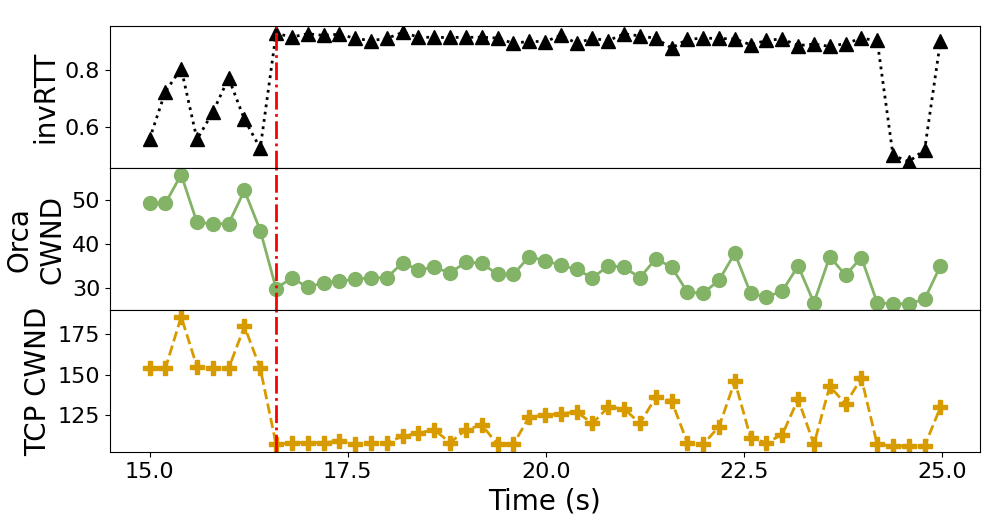}
    \caption{Trends of the observed invRTT ($=minRTT/RTT$), actual \cwnd action taken by Orca and the \cwnd suggested by TCP.}
    \label{fig:orca-critically-bad-state}
    \end{subfigure}

    \caption{\small{Orca entering critically bad states. \textbf{(a)} Orca's sending rate drops and remains low for a prolonged period while \sysname maintains its sending rate. \textbf{(b)} Orca's states for the region within the dashed lines in (a). Despite high invRTT (\textit{i.e.}, low queueing delays), TCP increases the \cwnd (>100) but Orca repeatedly reduces it (<50).
    }}
    \label{fig:orca-critically-bad}
 \vspace{-0.15in}\end{figure}

\noindent
\textbf{Observation \#2: Orca can enter bad states leading to poor utilization.}
Evaluating Orca on another synthetic trace (\Cref{fig:orca-critically-bad-trace}), 
we find that even without the impact of noise, Orca may enter into states with very low bandwidth utilization. We use high BDP
(bandwidth delay product) in this case.
To utilize the available bandwidth well, a controller should ideally meet the following property: increase (decrease) the \cwnd when experiencing low queueing (high) delays. Focusing on \cwnd increase behavior, \Cref{fig:orca-critically-bad-state} shows that at the start of the misbehavior by Orca (around 17s -- shown by the dashed vertical line), {\em even though TCP suggests a large \cwnd to fill up the available capacity quickly, Orca instead forces a significantly lower \cwnd and worsens the behavior}.
Furthermore, the result suggests that once Orca enters such bad states, it may fail to recover from them. 
Similarly, Orca can also rapidly increase \cwnd when it should not, leading to persistent delays, loss, and poor throughput (elided for brevity).

Such behaviors would be very unlikely had the Orca \lcc learner experienced the relevant input points during training -- if it were so, the learner would have seen a poor reward and ``corrected'' its action accordingly. However, \lccs cannot be expected to 
experience all possible values of inputs. %; in particular, Orca was  only trained on a finite (albeit large) number of traces and thus may not have seen queuing delays for all possible input states.
Such inevitable unexplored input spaces cause unpredictable decisions at run-time, leading to bad worst-case behaviors.

\subsection{\sysname: LCCs with Quantitative Certificates} \label{subsec:overview-certified-learning}

\lccs offer many real-world benefits but suffer from reliability issues of the above nature. To address this and improve operators' confidence  in \lccs, our solution, \sysname incorporates {\em symbolic worst-case analysis} of safety and robustness properties  
{\em into \lcc training}, such that, when trained, the \sysname \lcc provably meet operators' desired properties.
The result is shown in \Cref{fig:orca-critically-bad-trace,fig:orca-noises-trace}: \sysname produces controllers that are more robust to noise and achieve better utilization than Orca. 
%The end outcome of our approach is illustrated in \Cref{fig:orca-critically-bad-trace,fig:orca-noises-trace} -- C3 produces a controller more robust to noise and another that can ensure better utilization than Orca. \aditya{rephrased this paragraph}

\iffalse
\begin{figure}[t]
    \centering
    % First subfigure
    \begin{subfigure}{\linewidth}
        \centering
        \includegraphics[width=0.9\linewidth]{figures/motivation/robustness/throughput_bump500_24_12_c3.png}
        \caption{When the same amount of noise is applied, a \sysname model trained on a robustness property is more robust to noise, than Orca.}
        \label{fig:c3-vs-orca-robustness}
    \end{subfigure}
    % Second subfigure
    \begin{subfigure}{\linewidth}
        \centering
        \includegraphics[width=0.9\linewidth]{figures/motivation/fluctuation/throughput_bump500_96_48_c3.png}
    \caption{A \sysname model trained to avoid severe under-utilization gives better performance than Orca and avoids entering potentially bad states.}
    \label{fig:c3-vs-orca-fluctuations}
    \end{subfigure}

    \caption{\sysname yields models that are more robust to noise and recover from critically bad states.}
    \label{fig:c3-examples}
\end{figure}
\fi

We achieve this by leveraging Quantitative Certificate (\qc) feedback in the training loop.
A \qc contains a formal {\em proof} describing the extent to which a learned model satisfies a given property, as well as {\em quantitative feedback} that can be used to ``nudge'' learners towards property satisfaction. More precisely, the \qc's proof establishes that a property holds for \emph{all} possible inputs within a certain input region (e.g., all feature values in the vicinity of a feature measurement).
The feedback reflects the proportion of outputs from this region that provably satisfy the property, and the learning algorithm rewards LCCs if they expand this proportion. %We precisely define \qc in Section 4.3.
%includes a formal proof that describes how well a learned model satisfies a given property in the worst case, across {\em all possible} inputs within smaller, partitioned components of the input space.

For example, in an \lcc that takes network state (e.g., throughput, queuing delay) as input and outputs the \cwnd for the next time step, a property might state: "If queuing delay > $k \cdot RTT$ for a given period, the system should not increase \cwnd." A boolean proof, or certificate, for this fact would ensure that, for {\em any} network conditions matching the precondition -- {\em including those outside of the concrete conditions and environments the learner experiences during training} -- the model avoids increasing \cwnd. A
\qc refines this by partitioning the input space into smaller components and calculating the fractional volume of components provably satisfying the property (along with a proof that they do so). When this fraction reaches 100\%, the \qc aligns with the boolean certificate. %Further details are provided later in this section.

%Given a property, a \textit{quantitative certificate} is a proof over smaller components split from the input space, produced by a formal verifier, that quantifies to what extent a learned model satisfies a property in the worst case, i.e., for {\em all possible} inputs of each component. 
%For instance, for an \lcc that takes the network state (including throughput, queuing delay, etc.) as the input and outputs the \cwnd to be used for the next time step, a desirable property may be of the form: ``If queuing delay > $k \cdot RTT$ over a given period of time, the system should not increase the \cwnd.''
%Ideally, a boolean certificate wants to assure that {\em for all possible network conditions matching the precondition (queuing delay > $k \cdot RTT$ for a given period of time), the model's output avoids bad scenarios (does not increase \cwnd).} Intuitively, our quantitative certificates first break down the input space into fine-grained smaller components and then compute the fraction of input components that satisfy the property; more details are provided later in this section. The quantitative certificate is equivalent to the boolean certificate when the fraction of components satisfying the property is 100\%. 
%the model's output matches the post-condition (does not increase the \cwnd).

\begin{figure}[t]
    \centering
    \includegraphics[width=\linewidth]{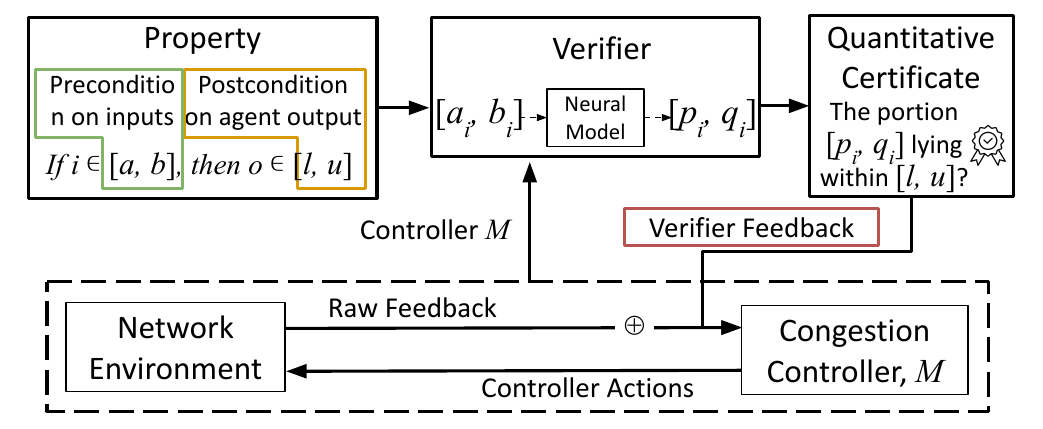}
    \caption{\small{\sysname Overview. Raw feedback is the training reward used by the base \lcc, and verifier feedback is the additional signal provided in \sysname used to construct the Quantitative Certificate. $[a_i, b_i]$ denotes one component of the input region $[a,b]$.}}
    \label{fig:solution-overview}
 \vspace{-0.15in}\end{figure}

\Cref{fig:solution-overview} overviews training in \sysname.
At each training step \footnote{Here, a step denotes one interaction of the \lcc with the environment.}, \sysname uses a verifier to 
compute a \certificate{}, then uses the \certificate{} to shape the reward of a {base} \lcc. In our specific implementation, the base \lcc is Orca.
%At each training step\footnote{Here, a step denotes one interaction of the \lcc with the environment.}, the verifier takes the current learned model (i.e., the Orca \lcc{}'s current weights) and the desirable property as inputs to compute a \certificate.
Here, 
given an input state range $[a, b]$ split into smaller components $\cup_i[a_i, b_i]$ $(\cap_i[a_i, b_i] = \emptyset)$, the verifier computes the intervals $\cup_i[p_i, q_i]$ overapproximating the set of outputs that the \lcc can possibly produce.
The verifier then checks whether the range of each model output component falls completely within the desired range $[l, u]$ defined in the property. 
We implement the verifier using abstract interpretation (\Cref{subsec:verifier}).

Each component's certification is boolean: a component either satisfies a property or not. However, since violation is more common than satisfaction,
%there are typically many more ways to fail a property than to pass it, 
a reward signal based on this for each component is \emph{sparse}, i.e., rarely positive. 
Learning from such sparse rewards is known to be hard \cite{eschmann2021reward}. 
Consequently, \sysname's computes an \emph{interval distance} between each output component range $[p_i, q_i]$ computed by the verifier and the desired output range $[l, u]$. This serves as \emph{\certificate{} feedback} (\Cref{subsec:certified-learning}) which is incorporated into \sysname's reward function alongside the "raw" reward used by the Orca \lcc for optimizing average-case performance. 
%within \sysname's reward function,  alongside the ``raw'' reward function that the underlying Orca \lcc uses to optimize average case congestion control performance.
While raw feedback captures empirical controller performance, verifier feedback provides a smooth measure of how close the controller is to satisfying the property in the worst case. By combining both, the learner is nudged toward good average-case performance {\em and} worst-case property satisfaction.
%While the raw feedback correlates with the empirical performance of the controller, the verifier feedback offers a smooth measure of how close the current controller is to provably conforming to the property. Thus, the learner is nudged toward good average-case performance {\em and} worst-case property satisfaction. 

We state a few salient features of the \sysname framework:
{\em First}, \lcc designers can customize the framework to incorporate specific properties, achieving a model tailored to their goals. This paper focuses on properties related to safety and robustness in \lccs (\Cref{subsec:properties}) for prototype. Yet, the primary contribution is integrating properties and their \certificate{}s into LCC training. %We do not mandate adherence to the specific properties discussed.
We do not impose strict adherence to the specific properties discussed; instead, users can customize properties as long as they follow the format outlined in \Cref{subsec:certified-rl-formulation}.
{\em Second}, \sysname decouples the \lcc backbone from the property, enabling its use with any \lcc beyond Orca. This flexibility extends to various learning algorithms, as the \certificate is not tied to a specific approach. %The certificate, provided alongside raw feedback, allows the agent to independently decide when and how to update its weights.
{\em Third}, the \certificate not only guides training reward construction but also measures the "strength" of property satisfaction independently. Thus, it can be computed for any controller to study how well it satisfies operators' desired properties.

\section{Background} \label{sec:bg}

This section provides a background on the Orca \lcc, abstract interpretation and its application to neural models.

\subsection{The Orca \lcc} \label{subsec:bg-orca}

Orca's RL formulation \cite{abbasloo2020classic}  uses a two-level control to determine a flow's \cwnd\ -- (i) fine-grained control performed by CUBIC~\cite{tcpCUBIC}, and (ii) coarse-grained control using a deep-RL agent.  At periodic time-steps, Orca's RL agent monitors real-time feedback from the network environment (\Cref{tab:orca-states}), including key metrics such as throughput and queuing delay -- that it uses to compute a new \cwnd and enforces this \cwnd at every coarse-grained control step (aligned with the monitoring period). Thus, Orca modulates TCP's behavior through ML-based suggestions, as follows:
\begin{equation}\label{eq:f-cwnd}
\small
\cwnd = f_{\cwnd}(a, \cwnd_{\tcp})=2^{2a} \times \cwnd_{\tcp}
\end{equation}
where $a \in [-1.0, 1.0]$ is the output of Orca's coarse-grained control, and $\cwnd_{\tcp}$ is the current TCP suggested $\cwnd$. 
During training, Orca employs a heuristic reward function based on the Power metric~\cite{power1981},
designed to increase as throughput rises or loss rate and queuing delay decrease:
\begin{equation}\label{eq:orca-reward}
\small
    R_\text{Orca} = \frac{R}{R_{\max}} = \left(\frac{\thr - \zeta \times \text{\emph{l}}}{\del'}\right) / \left(\frac{\thr_{\max}}{d_{\min}}\right) \\
\end{equation}
\begin{equation}
    \del' = 
    \begin{cases}
        d_{\min} & \left(d_{\min} \leq \del \leq \beta \times d_{\min}\right) \\
        \del & \text{o.w.}
    \end{cases}
\end{equation}
Here $\zeta$ is a coefficient for controlling the impact of loss rate compared to the throughput, $\thr_{\max}$ is the maximum throughput observed, and $\beta$ is a coefficient $>1$.

\begin{table}[t]
    \small
    \centering
    \begin{tabular}{c|l}
        \thr & The average throughput \\
        \hline
        \emph{l} & The average loss rate of packets \\
        \hline
        \del & The average queuing delay of packets \\
        \hline
        \emph{n} & The number of valid acknowledgement packets \\
        \hline
        \emph{m} & The time between current report and the last report \\
        \hline
        \emph{sRTT} & The smoothed RTT \\
    \end{tabular}
    \caption{O\small{bserved network states in Orca.} }
    \label{tab:orca-states}
\end{table}

\subsection{Abstract Interpretation} \label{subsec:verifier}

\begin{figure}[t]
    \centering
    \includegraphics[width=0.9\linewidth]{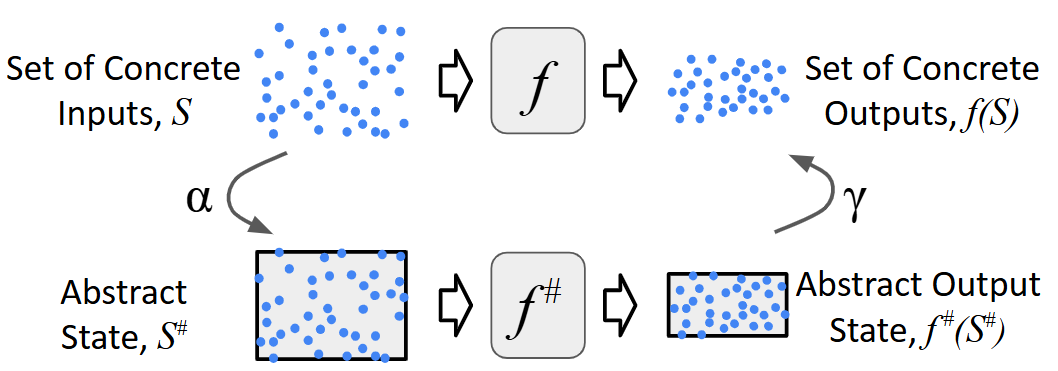}
    \caption{\small{Abstract Interpretation \textit{lifts} a concrete set of inputs to the abstract hyper-interval domain, and propagates the abstract state through the \textit{lifted} $f^{\#}$ function.}}
    \label{fig:abstract-interpretation}
\end{figure}

We build our quantitative certificates through \emph{Abstract Interpretation} \cite{cousot1977abstract}.
Here, one represents values --- e.g., the system state, controller action, and reward --- using symbolic representations, or \emph{abstractions}, in a predefined language (the \emph{abstract domain}).
For example, we can set our \emph{abstract states} to be hyperintervals with upper and lower bounds in each state space dimension. 
We denote abstract values with the superscript $\#$. For a set of concrete states $S$, $\alpha(S)$ (known as the \textit{abstraction function}) denotes the minimal-area abstract state containing $S$. For an abstract state $s^\#$,  $\conc(s^\#)$ (known as the \textit{concretization function}) gives the set of concrete states represented by $s^\#$ (\Cref{fig:abstract-interpretation}).

The core of abstract interpretation is the propagation of abstract states $s^\#$ through a function $f(s)$ that captures system dynamics.
For propagation, we assume that we have access to a map $f^\#(s^\#)$ that ``lifts'' $f$ to abstract states (\Cref{fig:abstract-interpretation}).
This function must satisfy the property
$\conc(f^\#(s^\#)) \supseteq \{f(s):  s \in \conc(s^\#)\}$, implying that $f^\#$ \emph{overapproximates} the behavior of $f$: while the abstract state $f^\#(s^\#)$ may include some states that are not reachable through the application of $f$ to states encoded by $s^\#$, it will \emph{at least} include {every} state that is reachable this way.
For simplicity of notation, we will use the notation $f(x)$ to represent functions when $x$ is concrete and their lifted counterparts when $x$ is abstract. For example, $\pi(s^\#)$ is short-hand for a function $\pi^\#(s^\#)$ satisfying $\conc(\pi^\#(s^\#)) \supseteq \{\pi(s): s \in \gamma(s^\#)\}$ for every abstract state $s^\#$.

% \diff{
\paragraph{Abstract State Propagation.}
In \sysname, we utilize the box domain \cite{mirman2018differentiable} for all the abstract states.
In the box domain, the abstraction function of a state with $m$ variables is represented by an $m$-dimensional box.
In formal terms, each abstract state is a pair $s^\# = {(b_c, b_e)}_{\#}$, where $b_c \in \Reals^m$ is the center of the box and $b_e \in \Reals^m_{\ge 0}$ represents the non-negative deviations.
% The $i$-th dimension of the concretization, $\gamma(s^\#)$, is given by interval
% \( [{(b_c)}_i - {(b_e)}_i, {(b_c)}_i + {(b_e)}_i] \).
% In other words, the abstract state $s^\# = {(b_c, b_e)}_{\#}$ for a box domain,
This abstract state $s^\#$ represents all concrete states for which the $i$-th dimension is in the interval \( [{(b_c)}_i - {(b_e)}_i, {(b_c)}_i + {(b_e)}_i] \).

We now showcase how the abstract state is propagated through the components of a neural network.
When certifying a neural model, we need to propagate the input interval (\Cref{fig:solution-overview}) through the model (denoted $\pi$), and any subsequent calculations (e.g., computation of $f_{\cwnd}$, defined in \Cref{eq:f-cwnd}).
This requires us to write abstract counterparts for the above functions.
As an example, below, we explain how $s^\#$ is propagated for the basic matrix multiplication operation in neural networks.
For a matrix multiplication function $f$ that multiplies the input $x \in \Reals^m$ by a fixed matrix $M \in \Reals^{m' \times m}$: $f(x) = M \cdot x.$
The abstraction function of $f$ is % given by:
$$
f^\#(s^\#) = ( M \cdot b_c, |M| \cdot b_e  )_\#,
$$
where $|M|$ is the element-wise absolute value of $M$. We present the abstract state propagation on `Add' and `ReLU' involved in the program and neural network certification below.

\textbf{Add.} 
The `Add' concrete function $f$ replaces the $i$-th element in the input vector $x \in \Reals^m$ with the sum of the $j$-th and $k$-th element:
$$
f(x) = (x_1, \dots, x_{i-1}, x_{j} + x_{k}, x_{i+1}, \dots x_m)^T.
$$
The abstraction function of $f$ is given by:
$$
f^\#(s^\#) = ( M \cdot b_c, M \cdot b_e )_\#
$$ where $M \in \Reals^{m \times m}$ is a matrix that allows $M \cdot b_c$ to replace the $i$-th element of $s^\#$ with the sum of the $j$-th and $k$-th element.

\textbf{ReLU.}
For a concrete element-wise $\textrm{ReLU}$ operation over $x \in \Reals^m$:
$$
\textrm{ReLU}(x) = (\text{max}(x_1, 0), \dots, \text{max}(x_m, 0))^T, 
$$
the abstraction function of $\textrm{ReLU}$ is given by:
\begin{align*}
\textrm{ReLU}^\#(s^\#) &= \left( \frac{\textrm{ReLU}(b_c + b_e) + \textrm{ReLU}(b_c - b_e)}{2}, \right. \\
&\quad \left. \frac{\textrm{ReLU}(b_c + b_e) - \textrm{ReLU}(b_c - b_e)}{2} \right)_\#.
\end{align*}
where $b_c + b_e$ and $b_c - b_e$ denotes the element-wise sum and element-wise subtraction between $b_c$ and $b_e$.
% }

\section{\sysname}\label{sec:certified-learning-framework}
In this section, we describe our learning problem (\Cref{subsec:certified-rl-formulation}), the properties \sysname targets (\Cref{subsec:properties}), the way we construct \certificates and incorporate them into training (\Cref{subsec:qc}). Finally, we present how \certificates can be used to monitor the quality of a deployed LCC (\Cref{subsec:qc-runtime}). 
%Then, we present our mechanism for incorporating \certificates into training (\Cref{subsec:certified-learning}). 
% reduce the overall optimization problem into a more tractable variant (\Cref{subsec:unconstrained-optimization}). 

\subsection{Problem}\label{subsec:certified-rl-formulation}
\noindent\textbf{RL Formulation.} \sysname uses Orca as its base \lcc. Following Orca, we formulate our learning problem in the general RL setting. For completeness and to aid the explanation of our approach, we provide the formal setting first. % of the setting first.

In RL, a learning task is modeled as a Markov Decision Process (MDP), defined as $\mdp = (\states, \actions, r, P, \states_0)$ where, $\states$ is a set of states; $\actions$ is a set of actions; $\states_0 $ is a distribution of initial states; $P(s' \mid s, a)$, for $s,s' \in \states$ and $a \in \actions$, is a probabilistic transition function; and $r(s, a)$ for $s\in \states, a \in  \actions$ is a real-valued reward function. 
A \emph{controller} (also called \emph{policy}) in $\mdp$ is a map $\pi$ from states $s$ to actions $a$.
% \footnote{In the implementation of \sysname, we only consider deterministic controllers.}.
A (finite) \emph{trajectory} $\tau$ under $\pi$ is a sequence  $s_0, a_0, s_1, a_1, \ldots$ such that $s_0 \sim \states_0$, each $a_i = \pi(s_i)$, and each $s_{i+1} \sim P(s' \mid s_i, a_i)$. We denote by $R(\tau) = \sum_i \gamma^i r(s_i, a_i)$ the discounted sum of rewards along a trajectory $\tau$, and by $R(\pi)$ the expected reward of trajectories unrolled under $\pi$.

For \sysname, 
$R$ is the raw reward function from Orca described in Eqn~\ref{eq:orca-reward}. We represent the state $s_t$ at time $t$ as the past $k$ steps' observation of the network states ($\langle o_t, o_{t-1}, \dots, o_{t-k}  \rangle$), $a_t$ as the adjustments to $\cwnd$, the policy $\pi$ as the congestion controller, and the transition as the network environment for a given link.
Our observations $o_i$ track the same set of monitoring statistics as in Orca (\Cref{tab:orca-states}). 

\paragraph{Constrained Optimization.} A \emph{property} in our setting is a constraint of the form $\phi(\pi, X, Y)$, such that $\forall \langle s_i, \dots, s_{i+k} \rangle \in X$, $a_{i+k} \notin Y$, where $X \subseteq \states^k$ is the precondition and $Y \subseteq \actions$ is the undesirable area within the action space in the postcondition of $\phi$ respectively. Our goal in this paper is to learn a controller that can \emph{maximize} a standard heuristic reward while certifiably satisfying the property.
That is, we expect to produce a \emph{certificate}, or \emph{proof}, that $\phi(\pi, X, Y)$ holds for the controller. We write $\pi \vdash_c \phi$ if $\pi$ satisfies $\phi$ for all $x \in X$ and $c$ is a certificate of this fact.
%Our base learning problem formalizes a setting in which the goal is to learn a controller that can \emph{maximize} a standard heuristic reward while certifiably satisfying a property. 
This learning problem can be written as 
\begin{equation}\label{eq:main-prob} 
\small
    (\pi^*, c) = arg\,max_{\pi \in \Pi} \expec_{\tau \sim (\mdp, \pi)} \left[ R(\tau)\right], \textrm{s.t.~~} \pi \vdash_c \phi
\end{equation}
where $R$ is the raw reward function of the MDP $\mdp$.

This problem is challenging in two ways: {\bf (1)} Achieving 100\% property satisfaction is hard for learning algorithms aiming to optimize a reward function, due to the high-dimensional search space. {\em Our solution} is to design a new optimization object, the quantitative certificate or \certificate, to {\em quantify} property satisfaction. We show how the \certificate can be integrated into learning to nudge a learner toward greater property satisfaction while also optimizing raw reward (\Cref{subsec:qc}); {\bf (2)} Even with \certificates, the above constrained optimization formulation leads to a complex search space where balancing the objective with the property-driven constraints can be challenging. {\em Our solution} is reducing the problem to an unconstrained one (\Cref{subsec:unconstrained-optimization}). Before digging into the details, we first introduce the properties we target below.
% 2).
% Since %Recall that in \Cref{sec:introduction} and \Cref{sec:motivation}, 
% 100\% property satisfaction is hard to achieve for learning algorithms \aditya{say why}, we design a new optimization object, the quantitative certificate or \certificate, to {\em quantify} property satisfaction. %\diff{As mentioned earlier, the \qc includes a region 
% %We partition the input space $X$ into $N$ subsets $\{X_n\}$, where $ \cap_{n} X_n = \emptyset \ \land \ \cup_{n} X_n = X$. 
% \diff{We expect our learning algorithm to produce, in addition to a controller $\pi$, a \emph{quantitative certificate}, that has two components 1). A "feedback" component that captures the fractional volume of the subset $X_\text{satisfy}$ of the overall input region $X$ ($X_\text{satisfy} \subseteq X$) that satisfies the property, and 2). A proof that $\phi(\pi, X_\text{satisfy}, Y)$ holds. We write $\pi \vdash_c \phi$ if $\pi$ satisfies $\phi$ for all $x \in X_\text{satisfy}$ and $c$ is a certificate of this fact.} %\aditya{add: The goal of our learning algorithm is to learn a $\pi$ such that the feedback $\rightarrow 1$.}

% This {\em constrained optimization problem} is difficult to solve - we present a tractable approach in Sections~\ref{subsec:certified-learning} and \ref{subsec:unconstrained-optimization}.

\subsection{Properties in \sysname} \label{subsec:properties}
% \aditya{the flow in this section is broken and it needs to be rewritten. first say that we are not coming up with prescriptions, simply borrowing from while. then talk about the four or five properties, and tie them to the observations in section 2.}

% \aditya{also, instead of called the 4-5 properties performance/robustness, we should number them. by given them names, we are incorrectly suggesting to reviewers that they are necessary for performance or robustness?}

Designers of \lcc{}s ideally want good average-case performance in terms of throughput and delays, while ensuring that the \lcc{} avoids bad behavior.
The reward metric used for training the backbone \lcc captures the average good performance. In contrast, to achieve the latter in \sysname, Congestion Control designers can rely on conventional wisdom -- detailed in the rich body of congestion control literature, or based on  their experiences of deploying and iterating over congestion control algorithms -- to craft properties that constraint the algorithm form taking bad actions. %such that bad actions are avoided.

We now present a set of five such properties P1-P5 detailed below. We study the effect of integrating them with \sysname in Section~\ref{sec:eval}.
Properties P1 and P2 correspond to network paths with shallow buffers and are derived from prior work on verifying congestion control (WhiRL~\cite{eliyahu2021verifying}).  P3--P5 are based on our observations in \Cref{sec:motivation}: P3 and P4 reflect desired properties for congestion control operating on paths with deep buffers; and P5 reflects desirable robustness to noise in measurements of path properties.
In defining these properties, we leverage a network transition model similar to WhiRL.\footnote{We follow WhiRL's overapproximation to transitions: The transition relation $T(x_1, x_2)$ as follows: for states $x_1$
and $x_2$, $T(x_1, x_2)$ is true if and only if the history vectors in $x_2$ are
precisely those in $x1$, shifted by one time step. In addition, the controller's output in $x1$ determines the state stored in
the newest entry in the bounded-length history of $x2$.} Note that these properties are not exhaustive, and don't necessarily reflect "ideal \lcc behavior". A congestion control designer can specify and customize properties as they gain experience with a particular \lcc deployment. % over the actions of an \lcc.

\begin{table*}[ht]
\small
    \centering
    \begin{tabular}{ c | p{14cm} }
        \textbf{Property} &   \multicolumn{1}{c}{\textbf{Formal Description}}\\
        \hline
        \textbf{P1} [Shallow, Good] & If at step $i$, the past $k$-steps' normalized queuing delays are in the range of $[0.0, q_{\del}^{\min} ]$, the past $k$-steps'  \\ 
         & normalized loss rates are $0.0$, and past $\Delta \cwnd$ $\leq 0.0$, then $\cwnd_i - \cwnd_{i-1}$ should $\geq 0$.\\
        \hline
        \textbf{P2} [Shallow, Bad] & If at step $i$, the past $k$-steps' normalized queuing delays are in the range $[0.0, q_{\del}^{\min} ]$, the past $k$-steps' \\
         & normalized loss rates are in $[p_\loss, 1.0]$, and past $\Delta \cwnd$ $\geq 0.0$, then $\cwnd_i - \cwnd_{i-1} $ should $\leq 0$. \\
        \hline
        \textbf{P3} [Deep, Good] & If at step $i$, the past $k$-steps' normalized queuing delays are in the range $[0.0, \; q_\del]$, packet losses are $0.0$, \\
        & and past $\Delta \cwnd$ $\leq 0.0$, then $\cwnd_i - \cwnd_{i-1} $ should $ \geq 0$. \\
        \hline
        \textbf{P4} [Deep, Bad] & i). If at step $i$, the past $k$-steps' normalized queuing delays are in the range $[p_\del, \; 1.0]$, and past $\Delta \cwnd$ $\geq 0.0$, then $\cwnd_i - \cwnd_{i-1} $ should $\leq 0$. ii). If at step $i$, the past $k$-steps' normalized queuing delays are in the range $[p_\del, \; 1.0]$, and for all past $\Delta \cwnd$ $\leq 0.0$, then $\cwnd_i - \cwnd_{i-1} $ should $\geq 0$. \\
        \hline
        \textbf{P5} [Robustness] & At step $i$, $\forall_{\eta \in [-\mu, \mu]} |\pi(s_i \times (1 + \eta)) - \pi(s_i)| \leq \pi(s_i)\times\epsilon$. Here, $\mu$ constrains the noise range on the observation, and $\epsilon$ allows the controller's output to fluctuate within a small range.
    \end{tabular}
    % \caption{\small{Formal Properties. In P5, $\mu$ is a parameter for constraining the noise range on the observation and $\epsilon$ allows the controller's output to fluctuate within a small range.}}
    \caption{{Formal description of properties in \sysname. The `step' in property description is the monitoring time period at which Orca interacts with the environment.}}
    \label{tab:formal-properties}
 % \vspace{-0.15in}
 \end{table*}

 We present the formal description of P1-5 in \Cref{tab:formal-properties}. Informal descriptions follow.

\textbf{Property 1.} {\em [Shallow buffer, good network condition]} When the congestion controller is sending on a path with shallow buffers (so the maximum queuing delay observed over a long period is small) and experiences no packet loss, the controller should eventually not decrease its \cwnd (note: a designer can also choose to increase current cwnd). % Formally, if at step $i$, the past $k$-steps' normalized queuing delays are in the range of $[0.0, {q_\del^\min} ]$, the past $k$-steps' normalized loss rates are $0.0$, and past $\Delta \cwnd$ $\leq 0.0$, then $\cwnd_i - \cwnd_{i-1}$ should $\geq 0$. 

\textbf{Property 2.} {\em [Shallow buffer, bad network condition]} When the congestion controller is sending on a path with shallow buffers and experiences high packet loss, the controller should eventually not increase its \cwnd. %Formally, if at step $i$, the past $k$-steps' normalized queuing delays are in the range $[0.0, {q_\del^\min}]$, the past $k$-steps' normalized loss rates are in $[p_\loss, 1.0]$, and past $\Delta \cwnd$ $\geq 0.0$, then $\cwnd_i - \cwnd_{i-1} $ should $\leq 0$. 

\textbf{Property 3.} {\em [Deep buffer, good network condition]} When the congestion controller is sending on path with deep buffers (maximum observed queueing delay is high) and experiences low queuing delays then the controller should eventually not decrease its \cwnd.
%Formally, if at step $i$, the past $k$-steps' normalized queuing delays are in the range $[0.0, \; q_\del]$, packet losses are $0.0$, and past $\Delta \cwnd$ $\leq 0.0$, then $\cwnd_i - \cwnd_{i-1} $ should $ \geq 0$. 

\textbf{Property 4.} {\em [Deep buffer, bad network condition]} When the congestion controller is sending on a path with deep buffers and experiences large queuing delay, we consider two sub-cases i). this flow caused congestion at the bottleneck queue, reflected in continued past non-decrease in \cwnd. ii). bottleneck congestion is due to other flows -- this flow already decreased our \cwnd. The controller should not keep increasing for case i) and not keep decreasing for case ii) to avoid further performance degradation for this flow. %Formally, for i), if at step $i$, the past $k$-steps' normalized queuing delays are in the range $[p_\del, \; 1.0]$, and past $\Delta \cwnd$ $\geq 0.0$, then $\cwnd_i - \cwnd_{i-1} $ should $\leq 0$. For ii), if at step $i$, the past $k$-steps' normalized queuing delays are in the range $[p_\del, \; 1.0]$, and for all past $\Delta \cwnd$ $\leq 0.0$, then $\cwnd_i - \cwnd_{i-1} $ should $\geq 0$.

\textbf{Property 5.} {\em [Noise Robustness]} Small {\em random noise} in the observed network state should not {\em drastically} change controller action (\Cref{sec:motivation}).\footnote{Note that such robustness properties are commonly desired of learned systems~\cite{tjeng2018evaluating,AI2,deeppoly-popl19}.
Further, because this property prevents drastic changes, it aligns with recent congestion controls that relay on delay gradients to make smooth \cwnd adjustments~\cite{timely}.}
%This property ensures that controller actions are robust to state perturbations arising from measurement errors (\Cref{sec:motivation}).
Ensuring robustness at each decision time-step also lowers the likelihood that the slight perturbations that inevitably occur during a single step snowball over future time steps and cause arbitrarily bad performance.

\subsection{Quantitative Certificate (\certificate) in Training}\label{subsec:qc}

We introduce a new optimization object, the quantitative certificate or \certificate, to {\em quantify} property satisfaction. \certificate has two pieces: (1) "Feedback" that approximates the fractional volume of the subset $X_\text{satisfy}$ of the overall input region $X$ prescribed in the property ($X_\text{satisfy} \subseteq X$) and (2) A "proof" that $\phi(\pi, X_\text{satisfy}, Y)$ holds. We rewrite $\pi \vdash_c \phi$ if $\pi$ satisfies $\phi$ for all $x \in X_\text{satisfy}$ and $c$ is the \certificate of this fact. \certificate reduces to the constraint $\phi(\pi, X, Y)$ in \Cref{subsec:certified-rl-formulation} when $X_\text{satisfy}=X$. We discuss these two pieces in turn next. %We first discuss the proof construction in \Cref{subsec:certificate-construction} and then present the training integration of the feedback in \Cref{subsec:certified-learning}.

\subsubsection{\certificate Proof Construction} \label{subsec:certificate-construction}

We now describe how the \certificate proofs are constructed in \sysname.
Constructing the proof requires us to partition the input space into components and verifying each component. For a given property, the relevant input space $X$ may be partitioned into $N$ disjoint subsets $\{X_n\}$, i.e., $ \cap_{n} X_n = \emptyset \ \land \ \cup_{n} X_n = X$. For simplicity, first consider that there is only one input component ($N=1$, $X_1 = X$). The process is similar for all components when $N>1$ and is outlined later. 

% \aditya{$k$ comes out of nowhere? Where are timesteps described. Can we link back?}
As described in \Cref{tab:formal-properties}, our properties are over the past $k$ steps and specifies precondition $X$ over some state observations $o$.
% Therefore, an abstract state in \sysname consists of $k$ intervals for each state dimension of interest, with each interval tracking the value of the dimension over one of the past $k$ time steps. 
Therefore, for a property $\phi(\pi, X, Y)$, the abstract state consists of $k$ intervals for each state dimension of interest, $o$, one interval each for the past $k$ time steps. 
Formally, the abstract state is represented as $s_{i}^\# = \langle o^\#_{{i-k}}, \dots, o^\#_{i} \rangle$.
Then we soundly propagate this abstract state through the controller to obtain symbolic output (action) $a^\#_i$, where $a^\#_{i} = \pi^\#(s^\#_{i})$.
By comparing $a^\#_{i}$ with $Y$, we obtain a proof of the property if $\gamma(a^\#_{i}) \nsubseteq Y$ ($\gamma$ is the concretization function from \Cref{subsec:verifier}).
Thus, the indicator function $\mathbbm{1}[\gamma(a^\#_{i}) \nsubseteq Y]$ is the \emph{formal proof} that a controller satisfies $\phi(\pi, X, Y)$. 

In the case of Orca, the output of $\pi(s)$ is used to modify the \cwnd (Eqn.~\ref{eq:f-cwnd}). Therefore, the abstract state propagation for Orca is given by:
% As the neural controller in Orca predicts a \emph{modification} to the value of \cwnd computed using \tcp, our abstract state propagation includes the following steps to allow $\cwnd_{i}^\#$ to soundly cover all the potential outputs from $s_{i}^\#$:
\begin{align}
\cwnd_{i}^\# = f_{\cwnd}^\#(\pi^\#(s_{i}^\#), \cwnd_{i}^{\tcp}),
\end{align}

\begin{table*}[ht]
\small
    \centering
    \begin{tabular}{ c |c | c}
        Property &  $X$ & $Y$\\
        \hline
        P1 &  $\{s | \forall j \in [i-k, i], o_j.\del \in [0.0, q_{\del}^{\min}] \land o_j.\loss \in [0.0, 0.0] \land \delta \cwnd_{j-1} \leq 0.0\}$ & $\{ \cwnd | \cwnd_{i} - \cwnd_{i-1} < 0\}$ \\
        \hline
        P2 & $\{s | \forall j \in [i-k, i], o_j.\del \in [0.0,q_{\del}^{\min}] \land o_j.\loss \in [p_\loss, 0.0] \land \delta \cwnd_{j-1} \geq 0.0\}$ & $\{ \cwnd | \cwnd_{i} - \cwnd_{i-1} > 0\}$ \\
        \hline
        P3 & $\{s | \forall j \in [i-k, i], o_j.\del \in [0.0, q_\del] \land o_j.\loss \in [0.0, 0.0] \land \delta \cwnd_{j-1} \leq 0.0\}$ & $\{ \cwnd | \cwnd_{i} - \cwnd_{i-1} < 0\}$ \\
        \hline
        P4 i) & $\{s | \forall j \in [i-k, i], o_j.\del \in [p_\del, 1.0] \land \delta \cwnd_{j-1} \geq 0.0\}$ & $\{ \cwnd | \cwnd_{i} - \cwnd_{i-1} > 0\}$ \\
        \hline
        P4 ii) & $\{s | \forall j \in [i-k, i], o_j.\del \in [p_\del, 1.0] \land \delta \cwnd_{j-1} \leq 0.0\}$ & $\{ \cwnd | \cwnd_{i} - \cwnd_{i-1} < 0\}$ \\
        \hline
        P5 & $\{s | s_{i} \times (1 + \eta), \eta \in [-\mu, \mu] \}$, $Y=\{ \cwnd| abs(\cwnd - \cwnd_{i}) / \cwnd_{i} > \epsilon\}$ & $\{ \cwnd| \frac{abs(\cwnd - \cwnd_{i})}{\cwnd_{i}} > \epsilon\}$
    \end{tabular}
    % \caption{\small{A formal description of $X$ and $Y$ in the proof components of \certificates for properties P1-5.}}
    \caption{{A formal description of $X$ and $Y$ in the proof components of \certificates for properties P1-5. $o_j$ represents the observation at step $j$ and $\delta \cwnd$ represents the change in \cwnd}}
    \label{tab:certificates}
    \vspace{-0.15in}
\end{table*}

\Cref{tab:certificates} shows the input components used for \certificate proofs  for properties P1-P5 (for $N=1$). %in \Cref{tab:certificates}. %defined above. 
For P1–4, at each step $i$, we track $\cwnd_{i-1}$ -- the \cwnd value from the previous step -- to compute the change $\delta \cwnd$.  P1-4's formal proof is $\mathbbm{1}[\Delta\cwnd^\#_i \nsubseteq Y]$, where $\Delta\cwnd^\#_i =\cwnd^\#_i - \cwnd_{i-1}$ is the action to be checked.
For P5, at each step $i$, we track $\cwnd_{i}$ given the observed states and the formal proof is $\mathbbm{1}[\cwnd{\textsc{change}}^\#_i \nsubseteq Y]$, where $\cwnd{\textsc{change}}^\#_i = (\cwnd^\#_i - \cwnd_i)/ \cwnd_{i}$, representing the \cwnd change fraction, is the action to be checked.

\noindent
{\bf Multiple input components.} We can extend the above proofs to multiple components ($N>1$) by certifying the input space $X = \cup_n X_n$.
\certificate's proof is then represented by $\mathbbm{1}[\land_n (\gamma(a^\#_{i_n}) \nsubseteq Y)]$, where $a^\#_{i_n}$ is the symbolic output representation from input component $X_n$.

\subsubsection{\certificate Feedback}\label{subsec:certified-learning}

We first present \certificate feedback construction. Then, we discuss its integration with the \sysname learning formulation.

% Even with an efficient mechanism for \certificate construction, our learning problem remains challenging for two reasons. First, the single component's property
% satisfaction is a sparse, discrete signal that is difficult to learn from. Second, \Cref{eq:main-prob} requires us to balance the objective of learning for better average-case performance (represented by $R$) with the goal of satisfying the property on worst-case inputs. To address these challenges, we further quantify the certificate satisfaction for training with a {\em \certificate function} and reduce the constrained problem into a tractable unconstrained problem. 
Given a property and $N$ input partitions $\{X_n\}$, we construct the \certificate feedback as the fractional volume of the abstract input for which the abstract output $a^\#_{i_n}$ satisfies the postcondition $Y_i$ (recall that $a^\#_{i_n}$ is $\Delta \cwnd^\#$ or $\cwnd\textsc{change}^\#$ for our properties of interest).
More concretely, the fractional volume of the abstract input $X_n$ is only included if its output region $a^\#_{i_n}$ is in the target area $\actions - Y$ (recall that $\actions$ is the action space and $Y$ is the undesired region). We define \qcf as $(\sum_{n\in N} \mathbbm{1}[a^\#_{i_n} \subseteq (\actions /\ Y)])/N$.

% between a target area (e.g. $\actions - Y = [y_l, y_u]$) and an output region (e.g. $\text{output}^\# = [o_l, o_u]$) is computed as:

% \diff{
To enable optimization, we {\em smoothen} $\certificate_\text{feedback}$ by considering the distance between $a^\#_i$ and $\actions /\ Y$ if $a^\#_i$ partially overlaps the target area:
% \certificate feedback as the fractional volume of the abstract output $\text{output}^\#_{i_n}$ from $X_n$ that satisfies the postcondition $Y_i$ (recall that $\text{output}^\#_{i_n}$ is $\Delta \cwnd^\#$ or $\cwnd\textsc{change}^\#$ for different properties). More generally, the fractional volume between a target area (e.g. $\actions - Y = [y_l, y_u]$) and an output region (e.g. $\text{output}^\# = [o_l, o_u]$) is computed as:
\begin{equation} %\small
    F(Y, a^\#_i) = \left \{ \begin{array}{lr}
       0,  &  \text{ if } a^\#_i \subseteq  Y\\
       1,  &  \hspace{-10pt} \text{ if } a^\#_i \subseteq \actions /\ Y \\ %y_l \le o_l \land o_u \le y_u\\
       \frac{\vol((\actions /\ Y) \cap a^\#_i)}{\vol(\text{output}^\#)}, & \text{other cases} \\
    \end{array} \right. 
\end{equation}
where \vol measures the volume of a set. For a given property and $N$ input partitions, we sum up all the fractional volume of each partition as $\hat\certificate_\text{feedback} = \frac{\sum_{n\in N}F(Y, \text{output}^{\#}_{n}))}{N}$. Intuitively, consider P5 (robustness): $\hat{\certificate}_{\text{feedback}}$ measures the proportion of the computed action range contained in the allowed noisy range.
% }

% We capture the input regions that satisfy the property by $X_\text{satisfy} = \cup_n X_n$, where $F(Y, \text{output}^{\#}_{n})$ is 1.0.

We now show how to leverage $\hat\certificate_\text{feedback}$ to guide learning. We construct a smooth  constraint satisfaction function $C(\pi, c, \phi)$ which reaches its maximum $C_{\max}$ when $\pi \vdash_c \phi$ (all components satisfy the property).
More precisely,
$C(\pi, c, \phi)$ $=$ $\expec_{\tau \sim (\mdp, \pi)} [$ $R_\text{verifier}(\tau) ]$, where $R_\text{verifier} = \sum_i \gamma^i r_\text{verifier}(s_i, a_i)$, for a reward function $r_\text{verifier}$ derived from the verifier. We use $\hat\certificate_\text{feedback}$ as the $r_\text{verifier}$ for single property.

When multiple properties, $P$, need to be satisfied, we average the different $\hat\certificate_\text{feedback}$ for the final reward:
\begin{equation}
    \small
r_\text{verifier}^{P} = \frac{\sum_{i \in P} \sum_{n\in N}F(Y^\#_{i}, \text{output}_{i_n}^{\#})}{N||P||}
\end{equation}
where $i$ are for different property cases. 

\subsubsection{From Constrained to Unconstrained Learning}\label{subsec:unconstrained-optimization}

Introducing the \certificate function converts \Cref{eq:main-prob} to
\begin{equation}\label{eq:modified-prob} \small
    (\pi^*, c) = arg\,max_{\pi \in \Pi} \expec_{\tau \sim (\mdp, \pi)} \left[ R(\tau)\right], \textrm{s.t.~~} C(\pi, c, \phi) = C_{\max}
\end{equation}

We can now use a known method \cite{le2019batch, freund1999adaptive} to reduce this problem to an unconstrained problem. The idea here is to \emph{convexify} the space $\Pi$ of controllers by allowing stochastic combinations of controllers, which expands $\Pi$ into its convex hull, allowing us to
% Given the convexity of the resulting controller space, 
rewrite \Cref{eq:main-prob} as 
\begin{equation}\hspace{-1.0em} \small
(\pi^*, c)=\min_{\hat{\lambda}} arg\,max_{\pi \in \hat{\pi}} \expec_{\tau \sim (\mdp, \pi)} \left[ R(\tau)\right] + \hat{\lambda} (C(\pi, c, \phi) - C_{max})
\end{equation}

Solving this optimally requires iterating over $\hat{\lambda}$. 
In practice, we treat $\hat{\lambda}$ as a hyperparameter balancing average-case performance and worst-case satisfaction, and omit convexification.
% simplify the problem by 
% assuming $\hat{\lambda}$ to be a hyperparameter 
% controlling the trade-off between average-case performance and the worst-case constraint satisfaction, 
% and omitting the convexification step.
$C_{\max}$ is treated as a constant and dropped during optimization.
% Also, $C_{\max}$ is treated as a constant to represent certification satisfaction that can be omitted during optimization. 
These simplifications leave us with the unconstrained optimization problem:
\begin{equation} \label{eq:unconstrained-learning-goal} \small
(\pi^*, c) = arg\,max_{\pi \in \Pi} \expec_{\tau \sim (\mdp, \pi)} \left[ (1- \lambda) R(\tau) + \lambda R_\text{verifier}(\tau)\right]
\end{equation}
where $\lambda \in \left[0, 1\right]$ is the said hyperparameter.
Intuitively, a small $\lambda$ tends to learn the original controller (e.g., Orca) and $\lambda \rightarrow 1.0$ makes the learning fully guided by worst-case property adherence. %\aditya{what lambda do we pick in our eval? Did we study sensitivity to this?} %\chenxi{We pick a $\lambda=0.25$ in Section 6 and present the sensitivity in appendix.} 
Eqn~\ref{eq:unconstrained-learning-goal} now has the form of a standard RL problem, which we can now apply to the underlying deep RL problem in Orca.

% We now use the underlying deep RL algorithm in Orca to solve this problem. %\aditya{do we want to quickly comment on how this extends to non-RL approaches?} \chenxi{I slightly prefer not. NSDI reviewers did not question this point. Also, the extension would not be straightforward. We already introduced the transition function following WhiRL. Any more complicated transition function can make the non-RL extension challenging.}

%\aditya{we can drop this whole paragraph. I don't think we need it any longer. Swarat?} \chenxi{I am fine with dropping. We already reinforce the QC in couple of places. }
In practice, neural networks seldom converge to the global optimum. Thus, while \sysname{} \emph{nudges} the learner toward property satisfaction, the resulting controller may not provably satisfy a property \emph{on all inputs}—especially if the designer chooses $\lambda \ll 1$. That said, the \certificate{} obtained at the end of training quantifies the extent of satisfaction and identifies input regions where the property holds. We denote the \qcf{} at convergence as \qcsat{}.

\subsection{Quantitative Certificate (\certificate) at Run Time}\label{subsec:qc-runtime}
\qcsat is generated alongside the model and serves as an additional system feature for online performance monitoring. We leverage feedback from \qcsat to adjust the controller's actions dynamically.

To support runtime adaptation, we incorporate a fallback mechanism guided by \qcsat. Given a learning-based congestion controller (LCC), we extract the \qcsat following the procedure in \ref{subsec:qc} prior to each decision step. The extracted \qcsat is then compared against a predefined threshold. The LCC's decision is applied only if the \qcsat meets the threshold; otherwise, the system falls back to TCP Cubic to safeguard property satisfaction.

\section{Implementation} \label{sec:implementation}
% \aditya{writing can be improved. Many details are similar to Orca - so put them together in one para and say "we borrow these from Orca". Separate the others out. say more about the parts that are different and unique to you.}

\noindent\textbf{Prototype and training setup.}
% for network state monitoring and for TCP Cubic to use \cwnd.% We modify the Linux Kernel (version 4.13) similar Orca for network state monitoring and for TCP Cubic to use the \cwnd computed by \sysname{} via shared memory.
% \noindent\textbf{Training.} 
We build the \sysname{} RL agent with Tensorflow \cite{abadi2016tensorflow} using the same neural controller architecture and agent learning algorithm (TD3) \cite{fujimoto2018addressing} as Orca. Training is distributed across 256 actors, each interacting with a distinct emulated network (via Mahimahi \cite{netravali2014mahimahi}) and one learner synchronizing neural parameters.

Each actor emulates a stable network link using Mahimahi, with bandwidth uniformly sampled between 6–192 Mbps and minimum RTT between 4–400 ms. The buffer size of this link is set to $0.5\mathrm{BDP}$, $5\mathrm{BDP}$, or $2\mathrm{BDP}$ based on the training property: shallow buffers (P1–2), deep buffers (P3–4), or robustness (P5), respectively. On each link, an actor runs a sender-receiver pair, with the sender continuously transmitting data. The sender’s congestion control algorithm is configurable and controlled by the actor.

We train all models with sixteen servers\footnote{Each server has 20-core Intel Xeon E5-2660 (2.60GHz) with 128GB RAM.} using a total of 256 cores for the actor pool; servers are connected via 100G links and another identical node is used for the learner.

\noindent\textbf{Verifier.}
We use Interval Bound Propagation (IBP) \cite{gowal2018effectiveness, zhang2019towards} as our base verification algorithm to extract the abstract output. To support IBP, we wrap the neural controller with Sonnet \cite{sonnet}, which can provide composable abstractions for each neural layer. As discussed in \Cref{sec:certified-learning-framework}, we partition the input into smaller components. After a careful analysis (details in supplementary material, \Cref{subsec:sensitivity}), we use $N=5$ for all \sysname training.

When representing the state input to the neural controller, we only abstract the variable of interest (e.g., queuing delay) and keep other variables (e.g., \# ACKs) as observed values. This allow us the freedom to explore different network conditions based on the concrete real-time states but also keep track of the worst-case effect from the variable of interest. %\aditya{not entirely sure I follow this.} \chenxi{We only abstract the variables in the precondition.}

% \begin{table}[t]
%     \centering
%     \small
%     \begin{tabular}{c|c}
%         Bandwidth & Minimum RTT\\
%         \hline
%         [6\emph{Mbps} - 192\emph{Mbps}] & [4\emph{ms} - 400\emph{ms}] \\ 
%     \end{tabular}
%     \caption{Training network environment characteristics.}
%     \label{tab:env-range}
%  \vspace{-0.15in}]
% \end{table}

\section{Evaluation}\label{sec:eval}
We evaluate \sysname by comparing against several baselines and answering the following questions for properties P1-5: 
\begin{enumerate}[label={\upshape\bfseries Q\arabic*:},] 
\setlength{\itemsep}{0pt plus 1pt}
    \item How does \sysname's \qcsat compare to Orca?
    \item How do \sysname's empirical performance measures -- throughput and delay -- compare to Orca and other congestion controllers?
    \item How well does \sysname perform in the wild?
    \item How does \qcsat-guided fallback influence runtime performance?
    \item Is \sysname fair and TCP-friendly? %(\Cref{subsec:sensitivity}) %Can \qcsat shed light on other congestion controllers?
    \item Is \sysname sensitive to system hyperparameters?
%\chenxi{TODO: add sensitivity here.}
\end{enumerate}
We study three \sysname models, one trained with {\em both} shallow buffer properties (P1 and P2), another with {\em both} deep buffer properties (P3 and P4), and one with robustness to noise property (P5). %For \#4, we focus on just the \sysname model trained with deep buffer properties. 
Our models are trained offline and are not tuned during deployment. 

% Our appendix offers additional analyses of \sysname, including %(A) How does \sysname perform with various hyperparameter settings? (\Cref{subsec:sensitivity}) 
% how does the training performance for \sysname compares against Orca (\Cref{app-subsec:training-perf}), and 
% ow much overhead is added by integrating the verifier in the training loop in \sysname? (\Cref{app-subsec:training-overhead})

% The appendix offers additional analyses of \sysname{}: (A) sensitivity to hyperparameter choices (\Cref{subsec:sensitivity}), (B) training performance compared to Orca (\Cref{app-subsec:training-perf}), and (C) training-time overhead from verifier integration (\Cref{app-subsec:training-overhead}).

The supplementary material provides additional details on \sysname{} such as: training performance vs. Orca (\Cref{app-subsec:training-perf}) and verifier-induced training overhead (\Cref{app-subsec:training-overhead}).
% and hyperparameter sensitivity (\Cref{subsec:sensitivity}).

\subsection{Experimental Setup} 
\label{sec:expt-setup}

\noindent\textbf{Baselines.}\quad We compare \sysname models against:
\newline\noindent\textbf{\em (i) Orca:} We train Orca for multiple rounds of 50k epochs each (as specified in the original paper~\cite{abbasloo2020classic}). We checkpointed the model after each round, and selected the checkpoint with the best performance for our evaluations in this section.
\newline\noindent\textbf{\em (ii) TCP Variants:} % We also evaluate three TCP variants: Cubic~\cite{tcpCUBIC}, Vegas~\cite{tcpVegas}, and BBR~\cite{tcpBBR}.
% Cubic treats packet loss as the main congestion signal and is used by Orca as its base congestion controller.
% Vegas instead uses delay as the main congestion signal.
% Finally, BBR combines observed delays and bandwidth estimation to make its decisions.

\noindent\textbf{Traces.}\quad We evaluate \sysname and the above baselines on 21 unseen traces, broadly of two categories:
\newline\noindent\textbf{\em (i) Synthetic Bandwidth Traces:} We construct 18 challenging synthetic traces (similar to the ones used in \Cref{subsec:motivation-examples}). These are similar to, but richer than, the traces used in SAGE \cite{yen2023computers} -- they incorporate more fine-grained bandwidth changes that SAGE traces do not capture. %, which captures typical link capacity change patterns. 
\newline\noindent\textbf{\em (ii) Cellular Traces:} We also use three cellular traces from \cite{winstein2013stochastic} with highly variable bandwidths from commercial LTE networks such as AT\&T, Verizon, and T-Mobile. 

\noindent\textbf{Real-world Deployment.}\quad We also set up a global testbed of VMs in nine different Azure regions, and use it to evaluate \sysname against baselines under complex real networks. Details about the setup are provided in \Cref{subsec:eval-realworld}.

\noindent\textbf{Methodology.}\quad We run a client that initiates a request to a server that responds by continuously sending data for the entire test duration. The congestion control scheme comes into play on the downlink from the server to the client.
The link between the client and the server is emulated using Mahimahi~\cite{netravali2015atc}.
We emulate each of the synthetic and real-world traces on the downlink, and test each scheme back-to-back for each trace 5 times and report the average.

At each time step of the evaluation when the agent takes the state and produces an action, we also compute a \certificate using the verifier with 50 components  equally splitting the input region corresponding to the property at hand.
We use a higher number of components than the training setup (\Cref{sec:implementation}) in order to precisely evaluate the derived \certificate.
%We use such certificates during evaluation to compare \sysname's achieved property satisfaction against Orca's (more on that below).

\noindent\textbf{Evaluation Metrics.}\quad %\diff{We use a variety of metrics to evaluate the performance of \sysname. 
First, we study the \qcsat, which is the \qcf at convergence (defined in \Cref{subsec:certified-learning}) to understand the property satisfaction of a congestion control scheme. A higher \qcsat reflects a stronger proof of satisfaction for the input region. Second, we evaluate the average bandwidth utilization and delay. %The average utilization captures how well a congestion control scheme is able to utilize the available bandwidth while delays denote the congestion controller's ability to maintain bounded queues. 
Ideally, we want \textit{high utilization and small, bounded delays}.
%The former is important because (as mentioned in \Cref{subsec:unconstrained-optimization}) in practice, \sysname{} may not
% necessarily arrive at a controller that yields 100\% \certificate satisfaction:

% \noindent\textbf{\em (i) \qcf:} 
% %Following \certificate, for each time step, FCC characterizes how many of the 50 components in the certificate satisfy the property being considered. This is slightly different the $r_\text{verifier}$ in training as we think fraction captures each components' satisfaction more clearly. We compute the mean and std of FCC over all time steps to obtain the FCC of a trace. \aditya{why did we choose this instead of the direct definition of QC's feedback? Can we simply replace this with QC feedback instead of calling it FCC?}

% \noindent\textbf{\em (ii) Average Utilization and Delay:} The average utilization captures how well a congestion control scheme is able to utilize the available bandwidth while delays denote the congestion controller's ability to maintain bounded queues. Ideally, we want \textit{high utilization and small, bounded delays}.

\noindent\textbf{\sysname Hyperparameters.}\quad We evaluate \sysname over a range of parameter choices for $\lambda$ and $N$. We choose the best performing \sysname model ($\lambda=0.25$, $N=5$) for the results that follow %in \Cref{subsec:eval-certificates,subsec:eval-empirical}, 
and perform a detailed sensitivity analysis in \Cref{subsec:sensitivity}. $q_{\del}^{\min}$ is $0.01$ for P1-P4. For buffer properties, we set  parameters $(q_{\del}=0.25, p_{\del}=0.75, p_{\loss}=0.75)$ and for robustness property we set $(\mu=0.05, \epsilon=0.01)$. We use $k=3$ steps. These parameters can be designer-customized. We follow Orca for all other hyperparameters in training.

\subsection{Evaluating \qcsat} \label{subsec:eval-certificates}
\begin{figure}[t]
    \centering
    \subfloat[Synthetic Traces]{%
        \includegraphics[width=0.49\linewidth]{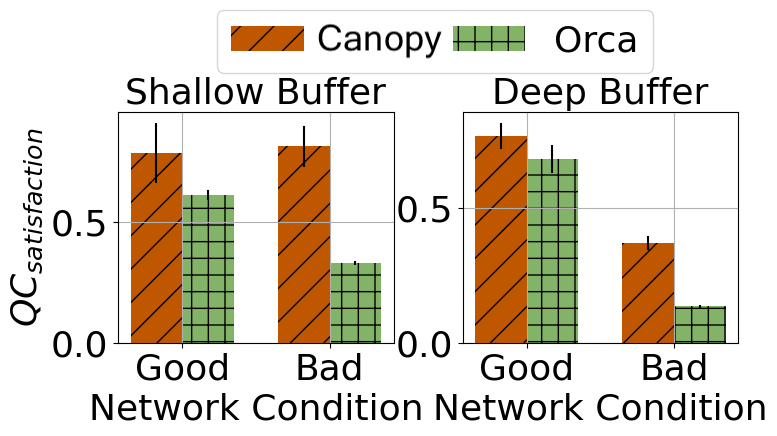}%
        \label{fig:certified-synthetic-small}
    }
    % Shallow Buffer
        % mean_c3_fcc_good_condition: 0.7794583333333334; mean_c3_fcc_bad_condition: 0.882375
        % mean_orca_fcc_good_condition: 0.6024166666666666; mean_orca_fcc_bad_condition: 0.3690833333333333
        % Deep Buffer
        % mean_c3_fcc_good_condition: 0.8236583333333334; mean_c3_fcc_bad_condition: 0.32581249999999995
        % mean_orca_fcc_good_condition: 0.6996249999999999; mean_orca_fcc_bad_condition: 0.11266666666666666
    \subfloat[Real World Traces]{%
        \includegraphics[width=0.49\linewidth]{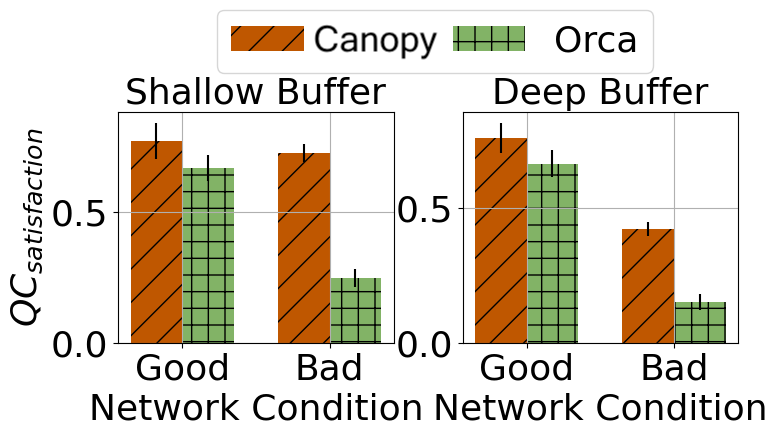}%
        \label{fig:certified-real-small}
    }
    \caption{\small{[Shallow Buffer \& Deep Buffer Properties] Mean and std of \qcsat on different categories of traces -- with trained buffer sizes ($0.5BDP$ for shallow buffer, $5BDP$ for deep buffer).}}
    \label{fig:certified-small}
\end{figure}

\begin{figure}[t]
    \centering
    \begin{subfigure}{0.49\linewidth}
        \centering
        \includegraphics[width=0.48\linewidth]{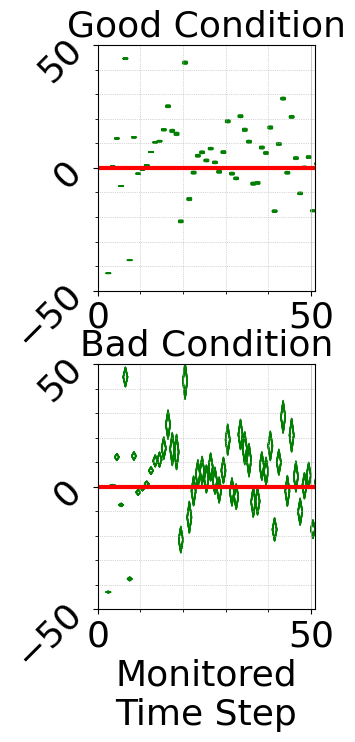}
        \includegraphics[width=0.48\linewidth]{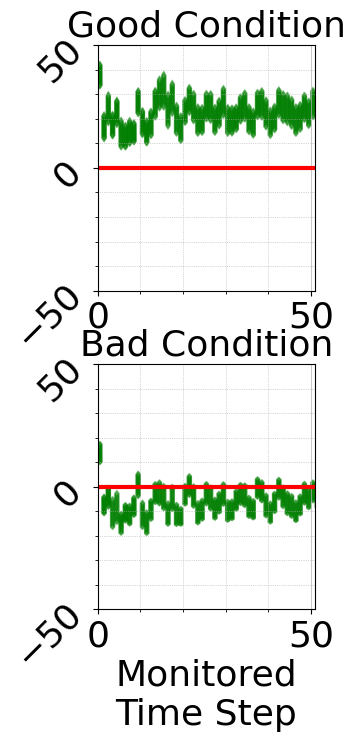}
        \caption{Trace 1: Orca (L), \sysname (R)}
        \label{fig:orca-certified-example-small}
    \end{subfigure}
    \begin{subfigure}{0.49\linewidth}
        \centering
       \includegraphics[width=0.48\linewidth]{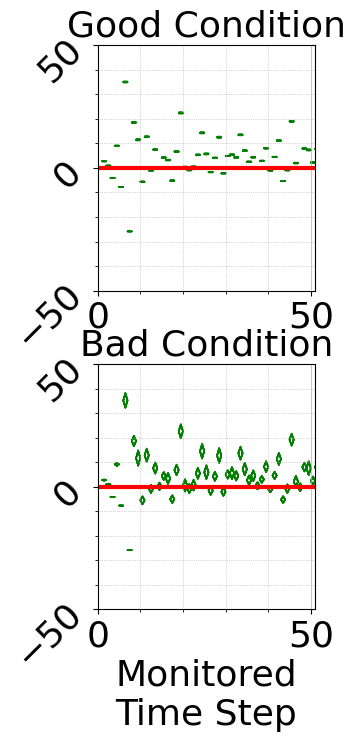}
        \includegraphics[width=0.48\linewidth]{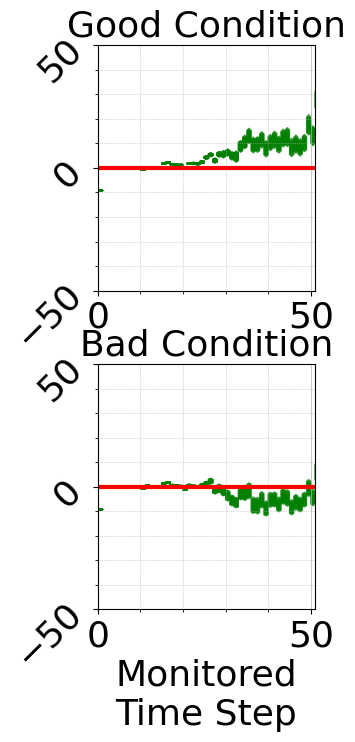}
        \label{fig:c3-certified-example-small}
        \caption{Trace 2: Orca (L), \sysname (R)}
    \end{subfigure}
    \caption{\small{[Shallow Buffer Property - y-axis: $\Delta \cwnd$] Certified components distribution of Orca (left) and \sysname (right) for 50 timesteps on two traces. Figures shows the output bounds of the \cwnd action ($\Delta \cwnd$) for each of the 50 components in the certificate (green-colored areas).
    For the good network condition case (top half in each trace), the property specifies $\Delta \cwnd \geq0$, i.e., colored regions above the red line are desirable.
    For the bad network condition case (bottom half in each trace), the property specifies $\Delta \cwnd \leq 0$, i.e., colored regions below the red line are desirable.
    }}
    \label{fig:certified-examples}
\end{figure}

We study the closeness of \sysname-trained controllers to property satisfaction. We also compare \sysname against Orca to understand the latter's property satisfaction.

\noindent\textbf{Shallow \& Deep Buffer Properties.}\quad
\Cref{fig:certified-small} shows the mean and standard deviation of \qcsat over all traces of a given type (synthetic/real), for two \sysname controllers, one trained with P1 and P2 (shallow) and another with P3 and P4 (deep). %(\Cref{subsec:properties}). 
\sysname can achieve a high 0.72-0.77 \qcsat over all traces for the shallow buffer properties; and 0.42-0.76 \qcsat for the deep buffer ones.
%Thus, on average, \sysname satisfies the shallow buffer property for 70\% of the input region.
In contrast, Orca can only provide 0.25 to 0.67 \qcsat for the shallow buffer case and 0.15 to 0.66 \qcsat for the deep buffer one. Thus, in practice, {\bf \sysname provides significantly higher worst-case satisfaction of the performance property.} 

We observe that \sysname{}'s \qcsat is poor in deep buffer scenarios. We believe this is due to the complexity of the P4 ("bad network") property, which consists of two sub-constraints. Since multiple properties exist within the deep buffer case, and all property rewards are treated equally during training, the learner tends to prioritize learning properties that are easier to satisfy, i.e., "deep buffer in good conditions". A designer can observe \sysname{} poor satisfaction in such cases, and as needed, re-weigh properties (P3 and P4 in this case) to achieve different outcomes. %Balancing property satisfaction across varying levels of complexity is an interesting direction for future work.

% This observation is also reflected in the \fcs numbers where \sysname achieves an \fcs of 0.50-0.66 (2.2-3.6$\times$ higher than Orca) for the large delay case of property and an \fcs of 0.62-0.71 (1.8-6.5$\times$ higher than Orca) for the small delay case.
% Intuitively, this implies that \sysname can provide a certificate with \textit{full} satisfaction of the property on $>50\%$ of the time steps.
% In particular, \sysname significantly outperforms Orca on real-world traces with large buffer sizes (see \Cref{fig:certified-real-large}), where it can meet the property on 6.5$\times$ more time steps than Orca.
% Thus \sysname can be expected to provide bounded delays in such cases (we will show this empirically below).

We provide a visual representation of what the \qcsat metrics imply in \Cref{fig:certified-examples} using two traces - for the depicted time slices, \sysname has better bounds on the \cwnd action for most components at any time (reflecting the high \qcsat metric). We find that regulating training with our properties helps the controller avoid continuously increasing or decreasing \cwnd too aggressively, which is ultimately reflected in the property satisfaction gain compared to Orca.

\begin{figure}[t]
    \centering
    \subfloat[Synthetic]{%
        \includegraphics[width=0.23\linewidth]{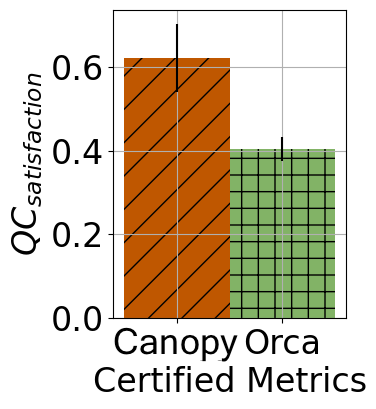}%
        \label{fig:certified-synthetic-robustness}
    }
    % synthetic
    % mean_c3_fcc: 0.6825; mean_c3_fcs: 0.435
    % mean_orca_fcc: 0.43; mean_orca_fcs: 0.005
    \subfloat[Real World]{%
        \includegraphics[width=0.23\linewidth]{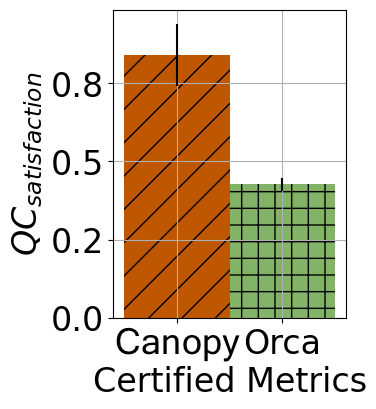}%
        \label{fig:certified-real-robustness}
    }
    % real world
    % mean_c3_fcc: 0.8075; mean_c3_fcs: 0.6950000000000001
    % mean_orca_fcc: 0.425; mean_orca_fcs: 0.042499999999999996
    \caption{\small{[Robustness Property] Mean and std of \qcsat on different categories of traces -- $2BDP$ buffer sizes.}}
    \label{fig:certified-robustness}
\end{figure}

\begin{figure}[t]
    \centering
    \begin{subfigure}{0.49\linewidth}
        \centering
        \includegraphics[width=0.48\linewidth]{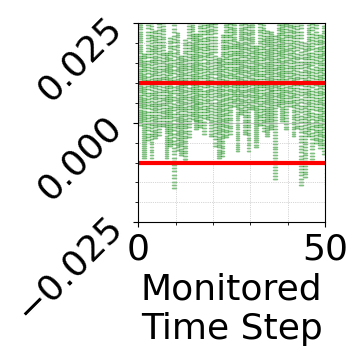}
        \includegraphics[width=0.48\linewidth]{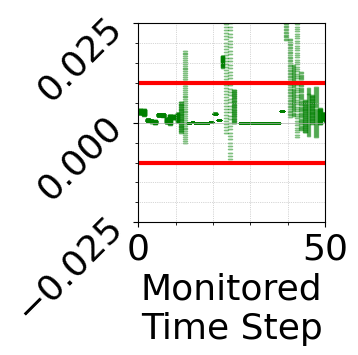}
        \caption{Trace 1: Orca (L), \sysname (R)}
        \label{fig:orca-certified-example-robustness}
    \end{subfigure}
    % Second subfigure
    \begin{subfigure}{0.49\linewidth}
        \centering
       \includegraphics[width=0.48\linewidth]{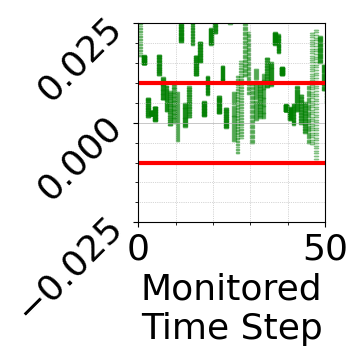}
        \includegraphics[width=0.48\linewidth]{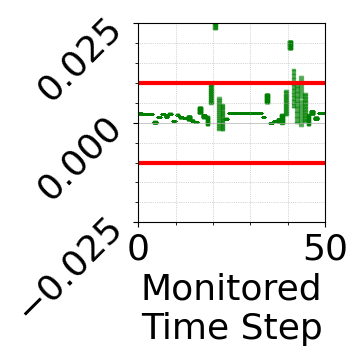}
        \label{fig:c3-certified-example-robustness}
        \caption{Trace 2: Orca (L), \sysname (R)}
    \end{subfigure}
    \caption{\small{[Robustness Property - y-axis: $\cwnd{\textsc{change}}$] Certified components distribution of Orca (left) and \sysname (right) for 50 time steps on two traces. The figures show the output bounds of the \cwnd change fraction ($\cwnd{\textsc{change}}$) for each of the 50 components in the certificate (represented in colored). The property specifies a target region for the \cwnd to fluctuate in, i.e., \cwnd{$\textsc{change}$} within the horizontal red lines at $y=\pm0.01$ is desirable.
    }}
    \label{fig:certified-examples-robustness}
\end{figure}

\noindent\textbf{Robustness Property (P5).} 
\Cref{fig:certified-robustness} presents the \qcsat results across two sets of traces for a controller trained with P5. \sysname achieves up to 0.81 \qcsat on real-world traces and 0.68 on synthetic traces -- thus, a majority of input regions with noise can formally satisfy the robustness property. In comparison, Orca has poor worst-case behavior: 
with an \qcsat of less than 0.05 when noise is introduced into the observed queuing delay. \Cref{fig:certified-examples-robustness} visually illustrates the $\cwnd{\textsc{change}}$ of two traces for Orca and \sysname\ -- \sysname better bounds the \cwnd change to the target area.

\subsection{Empirical Performance} \label{subsec:eval-empirical}
% We normalize the delays with respect to the minimum RTT for easier comparison.
We study three metrics of interest -- utilization, average delay, and p95 delay of our \sysname variants, and how they compare against Orca and key TCP baselines. We compute these metrics and evaluate these models using both synthetic and cellular traces.

\begin{figure}[t]
    \centering
    \subfloat[Synthetic Traces]{%
        \includegraphics[width=0.49\linewidth]{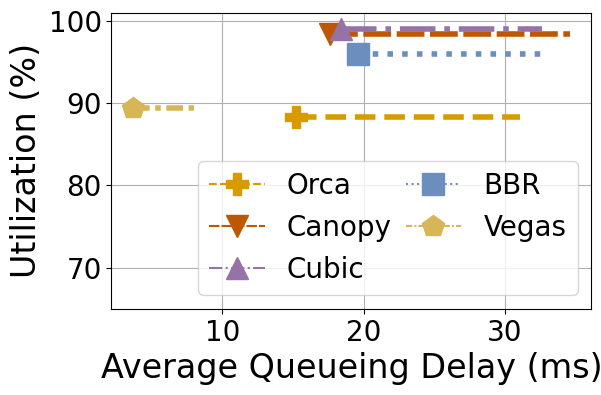}%
        \label{fig:empirical-synthetic-small}
    }
    \subfloat[Cellular Traces]{%
        \includegraphics[width=0.49\linewidth]{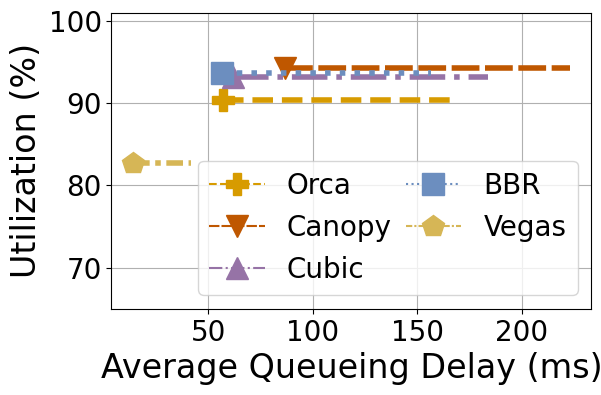}%
        \label{fig:empirical-real-small}
    }
    \caption{\small{[Shallow Buffer Property] Average throughput, normalized delay (icons) and p95 delay (end of lines). Top-left indicates better overall performance.}}
    \label{fig:thr-and-del-small}
\end{figure}
    
\begin{figure}[t]
    \centering
    \subfloat[Synthetic Traces]{%
        \includegraphics[width=0.49\linewidth]{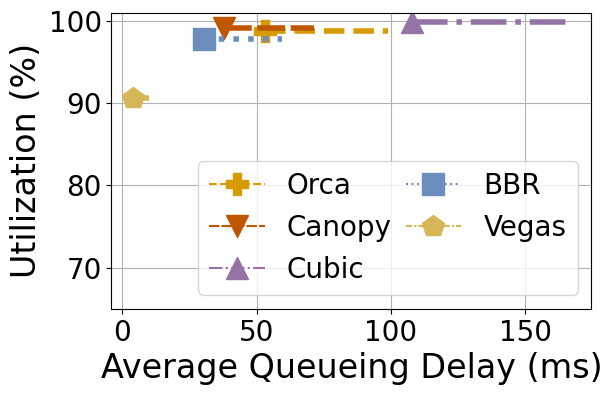}%
        \label{fig:empirical-synthetic-large}
    }
    \subfloat[Cellular Traces]{%
        \includegraphics[width=0.49\linewidth]{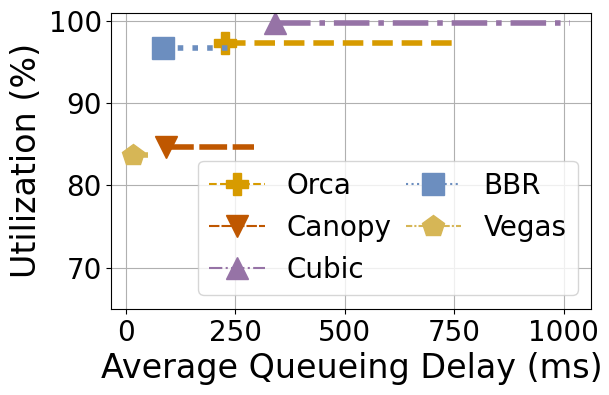}%
        \label{fig:empirical-real-large}
    }
    \caption{\small{[Deep Buffer Property] Average throughput, normalized delay (icons) and p95 delay (end of lines). Top-left is better.}}
    \label{fig:thr-and-del-large}
\end{figure}

\noindent\textbf{Shallow (P1-P2) \& Deep Buffer (P3-P4) Properties.} \quad 
\Cref{fig:thr-and-del-small,fig:thr-and-del-large} show the performance of our models and baselines on shallow buffers (=$1BDP$) and deep buffers (=$5BDP$), respectively.
We distill the results into key takeaways below.

\noindent\textit{\uline{Takeaway \#1: \sysname consistently achieves higher bandwidth utilization than Orca on shallow buffers.}}
\Cref{fig:thr-and-del-small} shows that compared to its base LCC (Orca), \textbf{\sysname improves average utilization by 4\% and 10\% on synthetic and cellular traces}, respectively, while incurring 11-33\% higher average p95 delays.
This is because property P1 (Table~\ref{tab:formal-properties}) encourages controllers to not continuously decrease \cwnd when delays are low -- resulting in higher utilization. \\

\noindent\textit{\uline{Takeaway \#2: \sysname consistently achieves lower p95 delays than Orca on deep-buffer networks.}}
On synthetic traces, \textbf{\sysname reduces average p95 delay by 28\% and slightly improves utilization (0.45\%)} over Orca.
On highly variable cellular traces, \textbf{it lowers average p95 delay by 61\% but incurs 12\% lower utilization}.
This drop in utilization may stem from conservative behavior: with deep buffers and high variability, delays can spike, and \sysname's P4 property discourages \cwnd increases under such conditions. This keeps delays low but sacrifices some bandwidth. \\

%, as evidenced by the performance gain from Orca to \sysname{}. However, we observe that \sysname{} does not perform well in deep buffer scenarios under poor network conditions, which aligns with our analysis in \Cref{subsec:eval-certificates}. We consider the investigation of this issue to enhance performance as future work.

%\aditya{main takeaway is not clear.}

\noindent\textit{\uline{Takeaway \#3: \sysname improves Orca's generalization,\\especially on shallow buffer links.}}
Orca performs well on deep buffers, evident by the near-perfect utilization for traces in \Cref{fig:thr-and-del-large}.
However, on shallow buffers (\Cref{fig:thr-and-del-small}) Orca underperforms - even falling behind TCP Vegas on synthetic traces (\Cref{fig:empirical-synthetic-small}).
This may stem from Orca being trained on $2BDP$ buffers — sufficient for learning deep-buffer behavior, while not having a similar opportunity to learn to utilize bandwidth well in shallow buffers.
% In particular, in shallow buffers, more losses may occur and to prevent concomitant throughput drops, it is important to avoid losses as soon as the delay increases.
\textbf{\sysname, though trained on the same data as Orca} ($2BDP$ buffers), learns to meet shallow-buffer properties and curbs \cwnd growth under high packet losses.
This results in \textbf{18-30\% fewer packets lost for \sysname}, driving the observed gains. \\

\noindent\textit{\uline{Takeaway \#4: \sysname has better delays than Cubic, and}} \\\textit{\uline{utilization similar to BBR.}}
{\sysname matches CUBIC's utilization on shallow buffers and improves delays on deep buffers.}
The deep buffer property (P4) used in \sysname might help in preventing Cubic's bufferbloat issue to a great extent -- providing 57-74\% smaller p95 delays for deep buffer links (\Cref{fig:empirical-synthetic-large,fig:empirical-real-large}).
\sysname also outperforms TCP Vegas on most traces, achieving 1-11\% higher absolute utilization than Vegas.
Against BBR, \sysname provides similar performance on shallow buffers, as shown in \Cref{fig:empirical-synthetic-small,fig:empirical-real-small}, where \sysname has 12-37\% smaller delays while achieving 0.92-0.98$\times$ bandwidth utilization. \\

\begin{figure}
    \centering
    \includegraphics[width=0.95\linewidth]{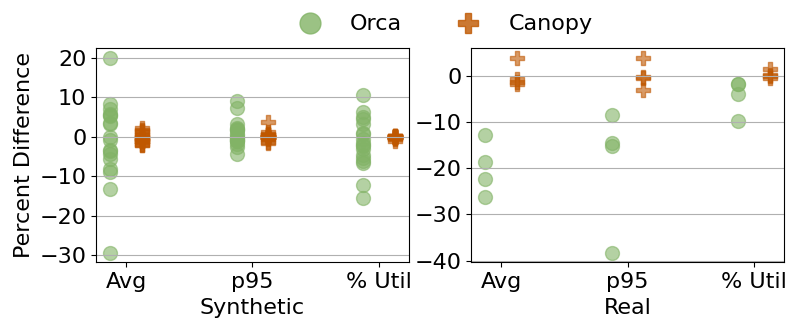}
    \caption{\small{[Robustness Property] Percentage change in the average delay (avg), p95 delay  (p95) and average utilization (\% Util) under 5\% random noise. Closer to zero implies more robust.}}
    \label{fig:robustness-thr-and-del}
\end{figure}

\noindent\textbf{Robustness Property (P5).}\quad
We apply a uniformly sampled noise within $[-5\%, 5\%]$ to the observed queueing delay for both Orca and \sysname. 
For each trace, we compute the percentage change in the three metrics (utilization, average delay, and p95 delay) when noise is added (\Cref{fig:robustness-thr-and-del}).
Overall, {\bf Orca is unpredictable and highly variable in the face of minor noise while \sysname trained with the robustness property is more robust.} 
While Orca can suffer up to 18\% drop in utilization, \sysname only sustains a maximum of 2\% drop while still maintaining 95\% utilization on average.
%\aditya{drop this?} Note that adding noise may cause Orca to inadvertently make a better decision at some time steps in some traces where it was performing poorly earlier, thus leading to cases where adding noise to Orca may improve its performance.

Overall, guiding training by identifying and avoiding undesirable behaviors can significantly enhance the learner’s performance.

\subsection{Performance in Real-world Deployments}\label{subsec:eval-realworld} 

\begin{figure}[t]
    \centering
    \subfloat[Intra-continental]{%
        \includegraphics[width=0.49\linewidth]{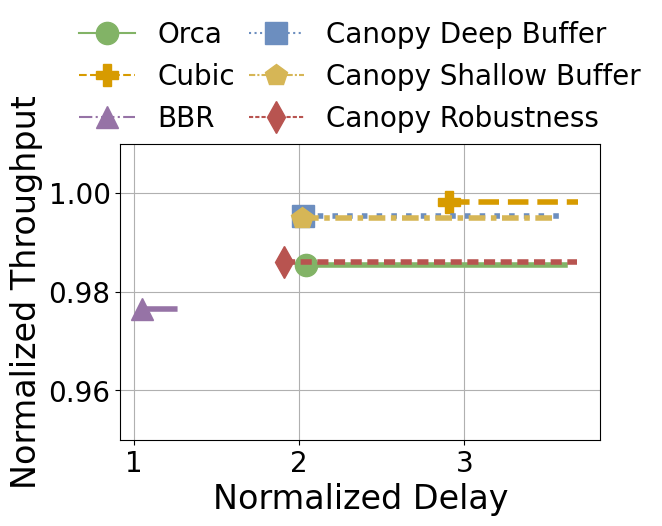}
        \label{fig:real-world-eval-america}
    }
    \subfloat[Inter-continental]{%
        \includegraphics[width=0.49\linewidth]{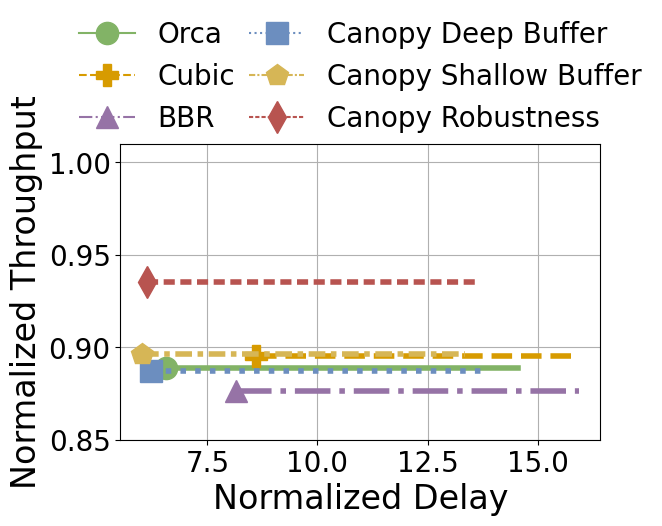}%
        \label{fig:real-world-eval-global}
    }
    \caption{\small{[Performance in the real world] Average throughput, normalized delay (icons) and p95 delay (end of lines) of \sysname variants and baselines. Top-left indicates better overall performance.}}
    \label{fig:real-world}
\end{figure}

We establish a sender in the Wisconsin cluster of Cloudlab (in the US) and run receiver clients on various Azure VMs.
We experiment with two categories of source-destination links: (a) Intra-continental links (client in EastUS, WestUS2, Canada, SouthCentralUS regions), and (b) Inter-continental links (client in Sweden, Australia, India, Brazil, South Africa).
Ping latencies between sender-receiver pairs range from 20ms to 237ms, representative of a wide range of network conditions.
We establish 30-second long flows between each source-destination pair, running each scheme five times.
For each source-destination pair, we then normalize the observed throughput and delay of a scheme by the maximum throughput and minimum delay, observed by any scheme, on this pair, respectively.
Figure~\ref{fig:real-world} shows the normalized throughput and normalized delay, for all schemes, averaged over all source-destination pairs in each category (intra- and inter-continental).
% We observe that for inter-continental links (Figure~\ref{fig:real-world-eval-global}) the \sysname robustness model outperforms all baselines, providing the best utilization (9.1\% higher than Orca, and 10.9\% higher than BBR). Additionally, the \sysname shallow buffer model has lesser delay than Orca, while preserving the bandwidth utilization.
% For the intra-continental links, both \sysname models - for deep and shallow buffers perform equally well and achieve 1.1\% higher throughput than Orca.
As observed in the emulated experiments in \Cref{subsec:eval-empirical}, \sysname's shallow buffer model provides higher bandwidths than Orca, and \sysname's deep buffer model provides lower delays than Orca.

\subsection{Runtime Performance with Fallback}

\begin{figure}[t]
    \centering
    \subfloat[Deep Buffer]{%
        \includegraphics[width=0.49\linewidth]{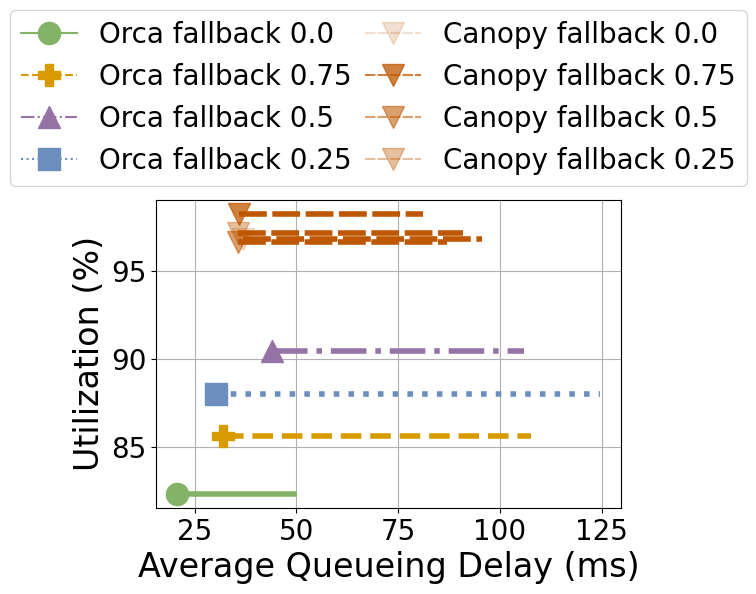}%
        \label{fig:fallback-deep}
    }
    \subfloat[Shallow Buffer]{%
        \includegraphics[width=0.49\linewidth]{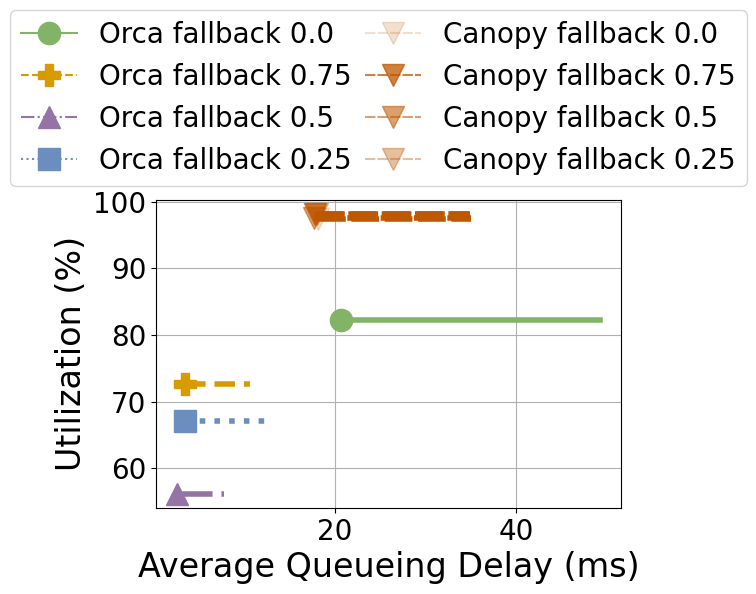}%
        \label{fig:fallback-shallow}
    }
    \caption{\small{[Runtime Performance with Fallback] Average throughput, normalized delay (icons) and p95 delay (end of lines). Top-left indicates better overall performance.}}
    \label{fig:thr-and-del-small}
\end{figure}

\begin{figure*}[t]
\begin{minipage}[b]{0.48\textwidth}
    \centering
    \subfloat[Friendliness on Shallow Buffers]{
        \includegraphics[width=0.46\textwidth]{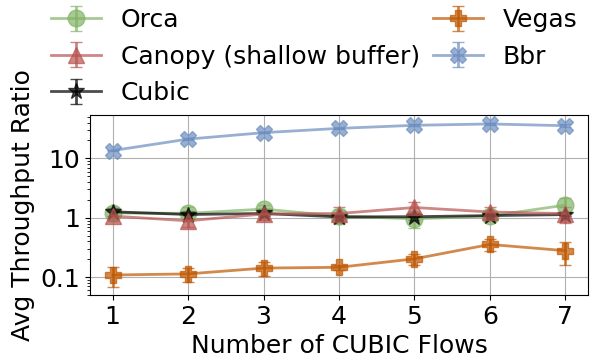}%
        \includegraphics[width=0.46\textwidth]{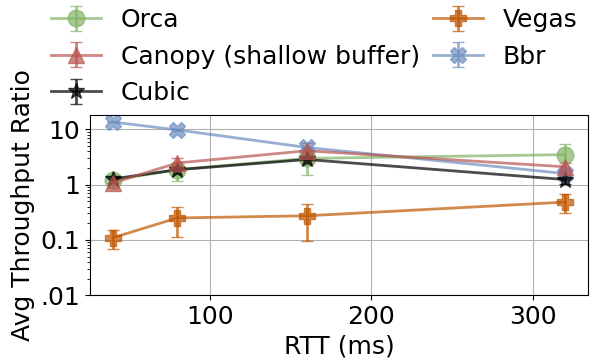}   
        \label{fig:eval-autoscaler-sn-normal}
    }\\
    \subfloat[Friendliness on Deep Buffers]{
        \includegraphics[width=0.46\textwidth]{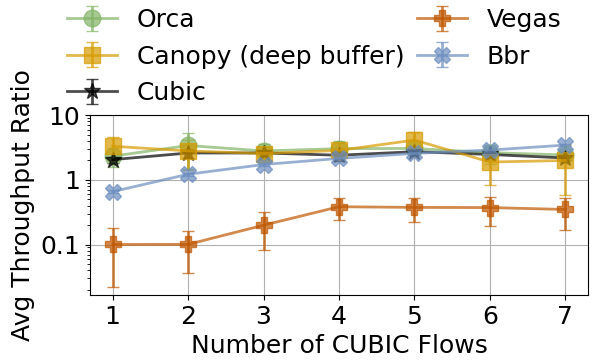}%
        \includegraphics[width=0.46\textwidth]{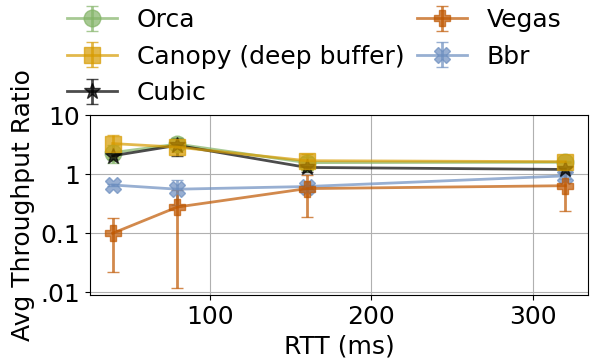}    
    }

    % \vspace{-0.1in}
    \caption{Throughput ratio to average of other CUBIC flows.}
    \label{fig:flowfriendliness}
\end{minipage}%
% \hfill
\begin{minipage}[b]{0.5\textwidth}
    \centering
    \includegraphics[width=0.48\textwidth]{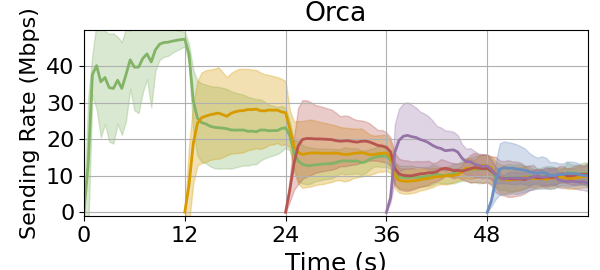}%
    \includegraphics[width=0.48\textwidth]{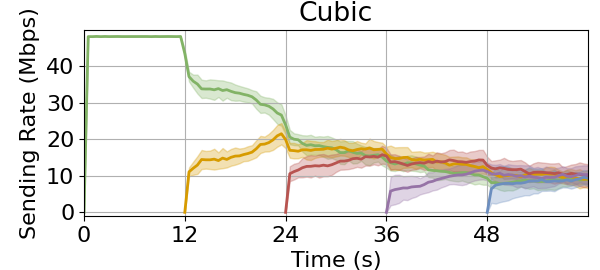} \\
    \includegraphics[width=0.48\textwidth]{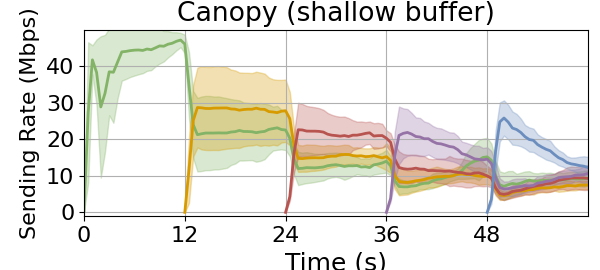}%
    \includegraphics[width=0.48\textwidth]{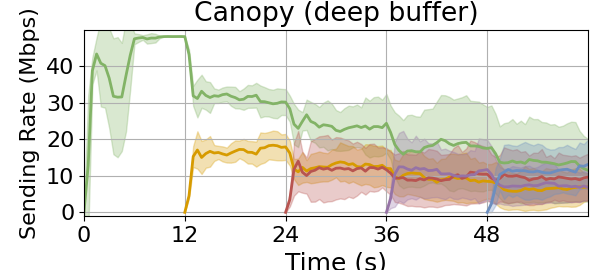} \\
    \includegraphics[width=0.48\textwidth]{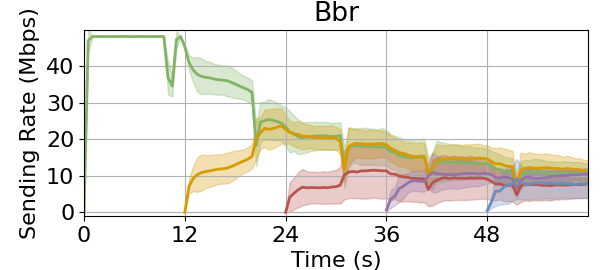}%
    \includegraphics[width=0.48\textwidth]{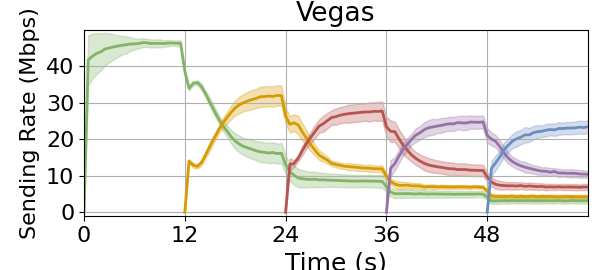}
    
    % \vspace{-0.1in}
    \caption{Convergence behavior of various schemes. }
    \label{fig:fairness_of_c3}
\end{minipage}%
\end{figure*}

\Cref{fig:fallback-deep} and \Cref{fig:fallback-shallow} present the results under varying \qcsat thresholds. At each decision step, the fallback mechanism consults \qcsat and defaults to TCP Cubic when the certificate does not meet the threshold. The results show that, with fallback, Orca achieves improved network utilization—indicating that \qcsat alone can serve as a useful runtime signal for property satisfaction monitoring. In contrast, the performance of \sysname remains largely unaffected by the fallback mechanism. We attribute this to \sysname being explicitly trained to satisfy the specified properties; as a result, it meets the threshold in most cases and rarely triggers fallback.

\subsection{Deployment Friendliness}
We evaluate whether the property-driven training in \sysname impacts the friendliness when running in the presence of other competing flows, and compare against Orca and the TCP baselines.
Since the properties were designed for specific network conditions, we evaluate the \sysname model trained with shallow buffer properties on 1BDP link and \sysname model trained with deep buffer properties on 5BDP link on an emulated Mahimahi setup (similar to the one used in \Cref{subsec:eval-empirical}).
For evaluating flow friendliness, we start an increasing number of competing CUBIC flows and measure the ratio of throughput achieved by the scheme under evaluation to the average throughput achieved by other CUBIC flows.
For evaluating RTT friendliness, we vary the propagation delay of the emulated link and start exactly one competing CUBIC flow.
We repeat these experiments 5 times for each scheme and report the mean and error bars in \Cref{fig:flowfriendliness}.
We observe that the behavior of the respective \sysname models is very close to the base LCC, Orca, which, in turn is close to the behavior of CUBIC as these schemes rely on CUBIC for fine-grained decisions.

\subsection{Fairness}
We evaluate the fairness of \sysname models in the presence of multiple homogeneous flows.
For each CC scheme under evaluation, we start a flow for 60s, introducing an additional flow using the same CC scheme every 12s. This was evaluated on a link with capacity of fixed 48Mbps bandwidth and 20ms delay, on a \textit{shallow} 1BDP link.
We repeated the experiment for each algorithm 10 times, and show the results in \Cref{fig:fairness_of_c3}. The shaded area shows the standard deviation of the runs.
Notably, the behavior of \sysname model trained on the shallow buffer property is very close to Orca's behavior, showcasing that the property does not worsen the fairness impact of the base LCC.
The \sysname model trained on the deep buffer property, however, takes longer to converge because the properties it was trained on, were designed for deep buffer links, thereby, impacting the overall convergence behavior. However, in the limit, as time progresses and with an increasing number of flows, the \sysname deep buffer model achieves convergence eventually (see \Cref{fig:fairness_of_c3}).  

% \diff{
\subsection{Sensitivity Analysis} \label{subsec:sensitivity}

\begin{figure}[t]
    \centering
    \includegraphics[width=0.75\linewidth]{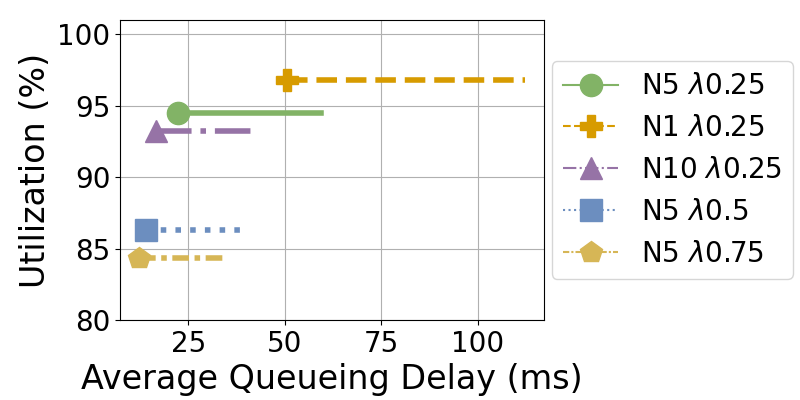}
    \caption{\small{[Shallow Buffer Property] Sensitivity of \sysname to different values of the hyperparameters. $\textsf{N}5\ \lambda0.25$ corresponds to the best-performing model used in \Cref{subsec:eval-certificates,subsec:eval-empirical}.}}
    \label{fig:sensitivity}
\end{figure}

% For brevity, we limit our analysis to the shallow buffer properties.
\Cref{fig:sensitivity} shows the performance of the \sysname model trained with the shallow buffer properties, for various parameters choices. We explain the insights below.

\noindent\textbf{Number of partitions.}\quad Note that we divide the input region into several partitions to improve the precision of certificates (\Cref{sec:implementation}).
A larger number of partitions implies a more precise certificate (and hence, more accurate \qcf) because the input region is divided into finer slices, thereby resulting in less over-approximation.
This is evident from \Cref{fig:sensitivity} where $N=1$ yields loose certificates and is hence unable to bound delays leading to 1.88$\times$ higher p95 delays. On the other hand, $N=10$ provides very accurate verifier feedback -- which helps it to achieve 27\% lower delays than $N=5$ but in the process it loses out on bandwidth utilization. Further, $N=10$ is also computationally expensive.
Overall, $N=5$ is found to be the best configuration balancing certificate precision and compute cost.

\noindent\textbf{Weight of the verifier reward.}\quad By substituting different values for $\lambda$ in \Cref{eq:unconstrained-learning-goal}, one can increase or decrease the weight given to the verifier reward during learning.
\Cref{fig:sensitivity} shows that increasing $\lambda$ from $0.25$ to $0.5$ and $0.75$ yields 32\% and 42\% smaller delays, respectively, but also results in 8.2\% and 10.1\% lower utilization as well.
The reason is that more weight to the verifier reward guides the learner to high property satisfaction (and hence, more bounded delays) but is unable to optimize for the raw reward.
\section{Related Work}

\noindent\textbf{Learning-based Congestion Controllers.} Several ML-based approaches have been proposed for congestion control. In addition to Orca \cite{abbasloo2020classic}, \cite{wei2021congestion} utilizes imitation learning, \cite{jay2019deep} introduces a deep RL-based controller, \cite{dong2018pcc} focuses on online learning, and \cite{winstein2013tcp} applies offline optimization. 
Additionally, \cite{orca-followup2023} explores data-driven congestion control design. However, none of these \lccs offer certified guarantees after training, and their neural networks can exhibit poor and unpredictable worst-case behavior similar to Orca's. 

\noindent\textbf{Verification for ML-based systems.}
Various machine learning techniques have been proposed for computer systems problems \cite{kanakis2022machine, maas2020taxonomy}, including resource allocation \cite{mishra2018caloree}, memory access prediction \cite{hashemi2018learning}, offline storage configuration recommendation \cite{klimovic2018selecta}, database query optimization \cite{kraska2021sagedb}, and storage placement optimization \cite{hao2020linnos, zhou2021learning}.
Formal verification is a powerful tool for assessing the worst-case performance of such systems,  as explored, for example, in  \cite{katz2021augmenting, dethise2021analyzing}.
Similarly, \cite{eliyahu2021verifying} verifies \lccs. All these approaches offer binary post-training verification feedback.

A recent system \cite{ouroboros-mlsys23} leverages verification tools during training to generate high-quality counterexamples to augment the training dataset. This approach produces a binary certificate at the end of training; also, like the above schemes,  there is no certificate feedback during training leaving no avenue to guide the learner toward property satisfaction. %. In addition, \cite{ouroboros-mlsys23}'s technique to identify input spaces to look for counterexamples applies only to supervised learning algorithms.

Another approach, \cite{gong2024zoom2net}, uses a linear programming solver to correct the output of the ML model. However, this method incurs high computational costs and is difficult to generalize to larger neural networks. In contrast, our framework uses properties solely to regulate the controller's bad behavior, eliminating the need for an optimal solution from the solver.

\noindent\textbf{Formal methods and learning.} There is a large literature on formal verification of neural networks \cite{pulina2011n,katz2017reluplex} and a growing body of work on certified learning techniques that integrate verification into the learning loop \cite{chaudhuri2014bridging,jothimurugan2019composable,mirman2018differentiable,yangsafe,yang2024certifiably}.
%races back to NaVer \cite{pulina2011n} and has since evolved with works like \cite{chaudhuri2014bridging} (which synthesizes a program following constraints), \cite{jothimurugan2019composable, jiang2021temporal} (which leverage temporal logic formulae to shape the RL reward), \cite{yangsafe} (which integrates verification into the training loop for mixed code-model settings), and \cite{yang2024certifiably} (which includes verification for model-based RL). % and \cite{anderson2020neurosymbolic} (which adds a learned shield to safeguard learning performance). 
However, these approaches have either considered restricted properties such as neural network robustness \cite{mirman2018differentiable}, or targeted small neurosymbolic programs \cite{yangsafe}. To our knowledge, we are the first to develop and apply a certified learning framework to the networked systems domain considering real-world operationally important properties.

\section{Concluding Remarks} \label{sec:discussion}
\sysname presents a novel approach to learning high-confidence congestion controllers by integrating symbolic worst-case analysis into the training loop. By leveraging quantitative certificates, \sysname guides training to improve worst-case property satisfaction while maintaining strong average-case performance. Our evaluation demonstrates that \sysname-trained models achieve significantly better worst-case guarantees than existing learned controllers, reducing delays while preserving bandwidth efficiency. These results suggest that incorporating formal analysis into learning can enable safer and more reliable RL-based congestion control in real-world deployments. {\em This work raises no ethical issues.}

\section*{Acknowledgements}
We thank the anonymous reviewers and our shepherd, Haoxian Chen, for their suggestions that improved this paper. This material is based upon work supported by the U.S. National Science Foundation (NSF) under Grant Number 2326576, and NSF awards CCF-2212559 and CCF-1918651. 

\balance
\bibliographystyle{plain}
\bibliography{ref}

% \newpage
\appendix
% \onecolumn
\section*{Appendix}
\section{Details on the Training of \sysname}
\subsection{Training Performance}\label{app-subsec:training-perf}
\begin{figure}[t]
    \centering
    \centering
    \includegraphics[width=0.8\columnwidth]{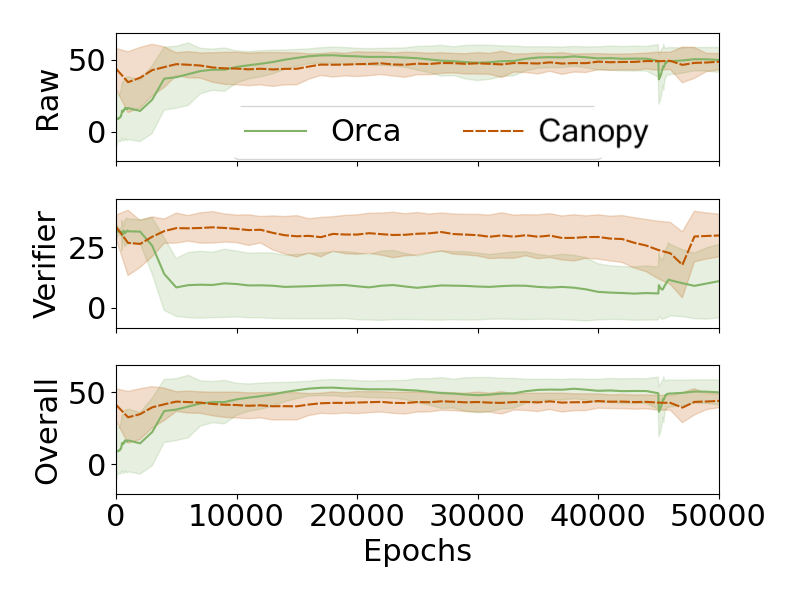}
    \caption{Training Curves of Orca and \sysname.
    }
    \label{fig:training}
\end{figure}

We choose the best-performing \sysname model trained on the shallow buffer properties -- $N=5, \lambda=0.25$ and compare the training performance for \sysname, against Orca.
We train Orca using the same procedure and training parameters as specified in the original paper~\cite{abbasloo2020classic}, picking the best checkpoint when trained on multiple training rounds.
We measure the raw reward, verifier reward, and the overall reward (as in \Cref{eq:unconstrained-learning-goal}) at each epoch and show the resulting training curve in \Cref{fig:training}.
As the training progresses, Orca's raw reward increases as expected. However, its verifier reward drops.
This indicates that just the raw reward is not enough to ensure high satisfaction of the property -- in fact, the decreasing verifier reward over epochs suggests that 
the Orca learner optimizes for the raw reward in a way that can reduce the satisfaction of the property! 
In \sysname, we can gain a better verifier reward, without significantly sacrificing the raw reward.

\subsection{Training Overhead}\label{app-subsec:training-overhead}

\begin{table}[th]
\small
    \centering
    \begin{tabular}{c|c|c|c}
    \multirow{2}{*}{Orca} & \multicolumn{3}{c}{\sysname\ - Shallow Buffer Property} \\ 
    \cline{2-4}   
       & $N$=1  & $N$=5  & $N$=10 \\
      \hline
      29.6 & 17.7 & 6.2 & 3.4 \\ 
    \end{tabular}
    \caption{Epoch rates.}
    \label{tab:training-overhead}
\end{table}

Unavoidably, verification introduces additional overhead. We measure the mean epoch rate (epochs per second) across 256 actors, as shown in \Cref{tab:training-overhead}. We vary the number of components ($N$) in the certificate (\Cref{sec:implementation}) to obtain the results. Each additional component requires another pass through the $\cwnd^{\#}$ computation.
In the performance property, we account for two delay constraints (\Cref{subsec:properties}), thus the verifier is invoked twice for each component. 
Therefore, the per epoch computational complexity is $O(\sysname) = (2N) \cdot O(\text{Verifier}) + O(\text{Orca})$. This results in the increased time for each epoch and reduced training throughput with increasing $N$.

\end{document}